\documentclass[runningheads]{llncs}
\usepackage{bm}
\usepackage{bbm}
\usepackage{siunitx} 
\usepackage{graphicx} 
\usepackage{amsmath} 
\usepackage[makeroom]{cancel}
\usepackage{mathtools}
\usepackage{algorithm}
\usepackage{booktabs}
\usepackage{amssymb}
\usepackage{url}
\usepackage{rotating}
\usepackage[noend]{algpseudocode}
\usepackage{appendix}

\usepackage[export]{adjustbox}
\graphicspath{ {Figures/} }
\usepackage{multirow}
\usepackage{pifont}
\usepackage{lineno,hyperref}
\usepackage{subcaption}
\usepackage{xcolor,colortbl}
\usepackage{caption}
\modulolinenumbers[5]
\usepackage{commath}
\graphicspath{ {figures/} }

\newcommand{\M}{\mathcal{M}}

\newcommand{\Xa}{X_{i}}

\newcommand{\x}{\vec{x}}

\newcommand{\Da}{D_{i}}

\newcommand{\LN}{L_\mathbb{N}}
\newcommand{\dn}{\delta L}
\newcommand{\hsiga}{\hat{\sigma}_i}
\newcommand{\hmua}{\hat{\mu}_i}
\newcommand{\hmud}{\hat{\mu}_d}
\newcommand{\hsigd}{\hat{\sigma}_d}

\newcommand{\Ytwo}{Y^2_i}
\newcommand{\Ytwon}{Y^2_n}

\newcommand{\Y}{\mathcal{Y}}
\newcommand{\Ya}{Y_{i}}

\newcommand{\Cands}{Cands}

\newcommand{\given}{\mid}

\algnewcommand\algorithmicinput{\textbf{Input:}}
\algnewcommand\INPUT{\item[\algorithmicinput]}
\algnewcommand\algorithmicoutput{\textbf{Output:}}
\algnewcommand\OUTPUT{\item[\algorithmicoutput]}
\newcommand{\ra}[1]{\renewcommand{\arraystretch}{#1}}
\DeclareMathOperator*{\argmin}{arg\,min}
\DeclareMathOperator*{\argmax}{arg\,max}
\DeclareMathOperator*{\loge}{\mbox{le}}

\renewcommand{\vec}[1]{\mathbf{#1}}

\definecolor{Silver}{rgb}{0.85,0.85,0.85}
\definecolor{Gray}{rgb}{0.5,0.5,0.5}

\newcolumntype{b}{>{\columncolor{Silver}}r}
\begin{document}
%
%
%

\title{Discovering outstanding subgroup lists\\for numeric targets using MDL}
\author{Hugo M. Proen\c{c}a\inst{1}\orcidID{0000-0001-7315-5925} \and
Peter Gr\"{u}nwald \inst{1,2}\orcidID{0000-0001-9832-9936} \and
Thomas B\"{a}ck \inst{1}\orcidID{0000-0001-6768-1478} \and
Matthijs van Leeuwen \inst{1}\orcidID{0000-0002-0510-3549}} 
\authorrunning{Hugo M. Proen\c{c}a et al.}
%
\institute{Leiden University, Netherlands \\
\email{\{h.manuel.proenca, T.H.W.Baeck, m.van.leeuwen\}@liacs.leidenuniv.nl} \\
\and
National Research Institute for Mathematics and
Computer Science in the Netherlands (CWI), Amsterdam, Netherlands \\
\email{Peter.Grunwald@cwi.nl}\\
} 

\maketitle              
\begin{abstract}
The task of subgroup discovery (SD) is to find interpretable descriptions of subsets of a dataset that stand out with respect to a target attribute. To address the problem of mining large numbers of redundant subgroups, subgroup set discovery (SSD) has been proposed. State-of-the-art SSD methods have their limitations though, as they typically heavily rely on heuristics and/or user-chosen hyperparameters. 

We propose a dispersion-aware problem formulation for subgroup set discovery that is based on the minimum description length (MDL) principle and subgroup lists. We argue that the best subgroup list is the one that best summarizes the data given the overall distribution of the target. We restrict our focus to a single numeric target variable and show that our formalization coincides with an existing quality measure when finding a single subgroup, but that---in addition---it allows to trade off subgroup quality with the complexity of the subgroup. We next propose SSD++, a heuristic algorithm for which we empirically demonstrate that it returns outstanding subgroup lists: non-redundant sets of compact subgroups that stand out by having strongly deviating means and small spread.

\end{abstract}

\section{Introduction}\label{section:intro}
\emph{Subgroup discovery} \cite{klosgen96sd,atzmueller2015subgroup} (SD) is the task of discovering subsets of the data that stand out with respect to a given target. It has a wide range of applications in many different domains \cite{proencca2018identifying}. For example, insurance companies could use it for fraud detection, where a found subgroup `$provider = $ HospitalX $ \wedge  $ $care = $ leg in cast $ \rightarrow average(claim) = \$2829.50$' might indicate that a certain health care provider claims much more for certain care than others.

Since its conception subgroup discovery has been developed for various types of data and targets, e.g., nominal, numeric~\cite{grosskreutz2019numeric}, and multi-label~\cite{van2010maximal} targets. In this paper we limit the scope to attribute-value data with a numeric target, i.e., each data point is a row with exactly one value for each attribute and a single, numeric target label, as is also considered in the regular regression setting.

\textbf{Related work.} Subgroup discovery traditionally focused on mining the top-k subgroups, based on their individual qualities. This approach has two major drawbacks: 1) its focus on quality measures that only take into account the centrality measure of the subgroup, such as the mean or median, and 2) the \emph{pattern explosion}, i.e., typically large amounts of redundant patterns are found. 

In response to the centrality problem associated with numeric targets, dispersion-aware (`spread-aware') measures---that allow for efficient mining of the top-k patterns---were proposed \cite{boley2017identifying}. These allow to directly take into account the dispersion of the subgroup target values when measuring its quality; therefore, they find more reliable subgroups. Nonetheless, these methods do not address the second drawback, i.e., the pattern explosion. 

To address this drawback, methods for \emph{subgroup set discovery} (SSD) have emerged. While SD aims on ranking the quality of subgroups regardless of how they cover the data together, SSD aims at finding good quality subgroups that together describe different regions of the data with minimum overlap between those regions. However, most of the SSD methods focus on binary target variables \cite{belfodil2019fssd,bosc2018anytime,lavravc2004subgroup}. For the setting with a numerical target variable, three approaches have been proposed:\\
\emph{1) Sequential covering}: CN2-SD~\cite{lavravc2004subgroup}, originally introduced for nominal targets, can be directly applied to numeric targets. The sequential covering idea is to iteratively find the subgroup having the highest quality, removing the data covered by that subgroup, and repeating this process until no further subgroups are found. This is virtually the same as mining a (ordered) list of subgroups and therefore closest to our approach.\\
\emph{2) Diverse Subgroup Set Discovery (DSSD)} \cite{van2012diverse}: DSSD uses a diverse beam search to find a non-redundant set of high-quality subgroups. It is based on a two-step approach that first mines a large pool of subgroups based on their individual qualities and then selects subgroups from that pool that maximize quality while penalizing for overlap. DSSD relies on tunable hyperparameters for the search and overlap penalization, which strongly influence the results. \\
\emph{3) Subjectively interesting Subgroup Discovery (SISD)} \cite{lijffijt2018subjectively}: This approach finds the subjectively most interesting subgroup with regard to the prior knowledge of the user, based on an information-theoretic framework for formalising subjective interestingness. By successively updating the prior knowledge based on the found subgroups, it iteratively mines a diverse set of subgroups that are also dispersion-aware.

Apart from the limitations already mentioned, all three approaches lack a \emph{global} formalization of the optimal set of subgroups for a given dataset and instead employ a sequential approach for which the stopping criteria, such as the total number of patterns to be found, need to be manually defined. 

\emph{Rule lists}, the model class to which subgroup lists belong to, were first proposed by Rivest~\cite{rivest1987learning} for the classification setting. Since then, many improvements have been proposed, from which the most related to our approach are: classification based on association rules or CBA \cite{ma1998integrating}; and both Bayesian\cite{yang2017scalable} and MDL\cite{proencca2020interpretable} formulations of probabilistic rule lists for classification. However, contrary to our formulation, all of these approaches have focused on prediction and nominal targets.

\textbf{Contributions.} We introduce a principled approach for dispersion-aware subgroup set discovery that builds on recent work \cite{vreeken2011krimp,proencca2020interpretable} that uses the minimum description length (MDL) principle \cite{rissanen78,grunwald2007minimum,grunwald2019minimum} for pattern-based modelling. The MDL principle states that the best model is the one that compresses the data and model best and is ideally suited for model selection tasks where the goal is to find succinct and descriptive models---such as is the case in subgroup discovery.

Informally, our MDL formulation of SSD aims at finding an ordered list of subgroups for numeric targets, that individually explain well different subsets of the dataset and together explain most of the data, while taking into account the spread of the target when measuring the quality of individual subgroups.

Our three main contributions are: \emph{1)} A formalization of subgroup set discovery for numeric targets using the MDL principle. To this end we devise a model class based on probabilistic rule lists. This probabilistic approach not only enables MDL-based model selection, naturally identifying compact subgroup lists, but also takes into account the dispersion (or spread) of the target value. By mining an ordered list of subgroups rather than an unordered set, we avoid the problem of a single instance being covered by multiple subgroups. This comes at the cost of slightly reduced interpretability, as the subgroups always need to be considered in order, but note that the still often-used sequential covering approach effectively identifies subgroup lists as well. \emph{2)} Derivations that show how our formalization relates to both an existing subgroup quality measure and Bayesian testing, and---based on these insights---a novel evaluation measure for subgroup lists. \emph{3)} SSD++, a heuristic algorithm that finds a set of non-redundant patterns according to our MDL-based problem formulation.

\begin{figure}[bth]\centering													
\ra{1.1}\small \begin{tabular}{@{}llrrrr@{}}										$s$\phantom{,,}	&	\textbf{description} of client bookings	&	$n$	&$	\hat{\mu}	$&$	\hat{\sigma}	$&	overlap	\\ 		\midrule
1	&\small	  month = $9$ \& customer\_type = Transient-Party	&$	22	$&$\phantom{,}	533	$&$	34	$&$\phantom{,}	-	$\\ 		
	&\small	  \& meal = Half Board \& country = GBR  \&  adults $\geq 2$	&		&		&		&		\\ 		\cmidrule(l){2-6}
2	&\small	 month $ \in [7,9]$\& market\_segment = Groups 	&$	29	$&$	336	$&$\phantom{,}	 \sim 0	$&$\phantom{,}	0 \%	$\\ 		
	&\small	\&  weekend\_nights = 1    \& distribution\_channel = Direct	&		&		&		&		\\ 		\cmidrule(l){2-6}
3	&\small	month = $9$ \& week\_nights =$4$ 	&$	16	$&$	343	$&$	3	$&$	0 \%	$\\ 		
	&\small	\&   distribution\_channel = Corporate 	&		&		&		&		\\ 		\cmidrule(l){2-6}
4	&\small	week\_nights = 0 \& deposit\_type = Refundable	&$	20	$&$	9	$&$	 \sim 0	$&$	0 \%	$\\ 		
	&\small	 \& repeated\_guest = no \& adults$ \geq 2$	&		&		&		&		\\ 		\midrule
\multicolumn{2}{l}{dataset overall distribution}			&$	18 \:550^*	$&$	92	$&$	99	$&$	-	$\\ 		\bottomrule
\end{tabular}													
\caption{First $4$ subgroups of a subgroup list obtained by SSD++ on the \emph{Hotel booking} dataset with target \emph{lead days}---number of days in advance the bookings were done (this case study is discussed in Section~\ref{section:casestudy}). \emph{Description} contains information regarding client bookings, $n$ the number of instances covered, $\hat{\mu}$ and $\hat{\sigma}$ are the mean and standard deviation in days, and \emph{overlap} is the percentage of the subgroup description that is covered by subgroups that come before in the list, i.e., how independently can the subgroups be interpreted. The last line represents the dataset overall probability distribution. $^*$ The $n$ of the dataset is the total number of instances in the dataset.}\label{fig:hotelnorm_application}
\end{figure}

\textbf{Example.} To illustrate how our MDL-based problem formulation naturally defines a succinct and non-redundant set of subgroups for a given dataset, without the need to define the desired diversity or number of patterns in advance, we show an example subgroup list as obtained by our approach on the \emph{Hotel booking} dataset (see Figure~\ref{fig:hotelnorm_application} for the details and in depth explanation in Section~\ref{section:casestudy}). Our method identifies a detailed list of booking descriptions from which we show here the first four subgroups, each consisting of a short description that clearly represent different sub-populations of the data, i.e., different types of client bookings.

\section{Subgroup Discovery with Numeric Targets}	
\label{section:sd}

Consider a dataset $D = (X, Y) = \{(\x_1,y_1),(\x_2,y_2),...,(\x_n,y_n)\}$. Each example $(\x_i,y_i)$ is composed of a numeric target value $y_i$ and an instance of values of the explanatory variables $\x_i = (x_{i1},x_{i2},...,x_{ik})$. Each instance value $x_{ij}$ is associated to variable $v_j$ and the total number of values in an instance is $k=|V|$ values, one for each variable $v_j$ in $V$, which represents the set of all explanatory variables present in $X$. The domain of a variable $v_j$, denoted $\mathcal{X}_{j}$, can be one of three types: numeric, binary, or nominal (with $>2$ values). $Y$ is a vector of values $y_i$ of the numeric target variable with domain $\Y = \mathbb{R}$.

\smallskip
\noindent \textbf{Subgroups}. A subgroup, denoted by $s$, consists of a \emph{description} (also intent) that defines a \emph{cover} (also extent), i.e., a subset of dataset $D$. 

\emph{Subgroup description}. A description $a$ is a Boolean function over all explanatory variables $V$. Formally, it is a function $a : \mathcal{X}_{1} \times \cdots \mathcal{X}_{|V|} \mapsto \{false,true\}$. In our case, a description $a$ is a conjunction of conditions on $V$, each specifying a specific value or interval on a variable. The domain of possible conditions depends on the type of a variable: numeric variables support \emph{greater and less than} $\{\geq, \leq\}$; binary and categorical support \emph{equal to} $\{ =\}$. The size of a pattern $a$, denoted $|a|$, is the number of variables it contains. In Figure~\ref{fig:hotelnorm_application}, subgroup 1 has description of size $|a| = 5$, where two of those conditions are $\{\mbox{meal} = \mbox{Half Board} \}$ and $\{\mbox{adult} \geq 2\}$; on a categorical and a numerical variable, respectively.

\emph{Subgroup cover}. The cover is the bag of instances from $D$ where the subgroup description holds true. Formally, it is defined by $D_{a}= \{(\vec{x},y) \in D \given a \sqsubseteq \vec{x} \}$, where we use $a \sqsubseteq \vec{x}$ to denote $a(\vec{x}) = true$. Further, let $|D_a|$ denote the coverage of the subgroup, i.e., the number of instances it covers. 

\emph{Interpretation as probabilistic rule}. As $D_a$ encompasses both the explanatory variables and the target variable, the effect of $a$ on the target variable can be interpreted as a probabilistic rule $a \mapsto \hat{f}_a(Y)$ that associates the antecedent $a$ to its corresponding target values in $Y$ through the empirical distribution of their values $\hat{f}_a(y)$. Note that in general $\hat{f}_a(Y)$ can be described by a statistical model and corresponding statistics $\hat{\Theta}$, e.g., a normal distribution $\mathcal{N}(\mu,\sigma)$ with estimated mean $\hat{\mu}$ and standard deviation $\hat{\sigma}$. 

Revisiting the subgroup list in Figure~\ref{fig:hotelnorm_application}, the description and corresponding statistics for the third subgroup are $a =$ $\{$month $= 9$ \& week\_nights $= 4$ \&  distribution\_channel $=$ Corporate$\}$ and $\hat{\Theta}_a = \{ \hat{\mu} = 343; \hat{\sigma} = 3\}$, respectively, and together represent the following rule:
\begin{equation*}
\textsc{ if }    a  \sqsubseteq  \x   \textsc{ then }  \mbox{lead time} \sim \mathcal{N}(\mu = 343; \sigma = 3 ) \\
\end{equation*}
where $\mathcal{N}(\mu = 343; \sigma = 3 )$ is the probability density function of a normal distribution. 

\smallskip
\textbf{Quality measures}.
To assess the quality (or interestingness) of a subgroup description $a$, a measure that scores subsets $D_a$ needs to be chosen. The measures used vary depending on the target and task \cite{atzmueller2015subgroup}, but for a numeric target it usually has two components: 1) representativeness of the subgroup in the data, based on coverage $|D_a|$; and 2) a function of the difference between a statistic of the empirical target distribution of the pattern, $\hat{f}_a(Y)$, and the overall empirical target distribution of the dataset, $\hat{f}_d(Y)$. The latter corresponds to the statistics estimated over the whole data, e.g., in Figure~\ref{fig:hotelnorm_application} it is $\hat{\Theta}_d = \{ \hat{\mu} = 92; \hat{\sigma} = 99 \}$ and it is estimated over all $18\:550$ instances of the dataset.

The general form of a quality measure to be maximized is
\begin{equation}\label{eq:generalmeasure}
q(a) = |D_a|^\alpha g(\hat{f}_a(Y),\hat{f}_d(Y)), \; \alpha \in [0,1],
\end{equation}

where $\alpha$ allows to control the trade-off between coverage and the difference of the distributions, and $g(\hat{f}_a(y),\hat{f}_d(y))$ is a function that measures how different the subgroup and dataset distributions are. The most adopted quality measure is the Weighted Relative Accuracy (WRAcc)\cite{atzmueller2015subgroup}, with $\alpha=1$ and $g(\hat{f}_a(Y),\hat{f}_d(Y)) = \hat{\mu}_a-\hat{\mu}_d$ (the difference between averages of subgroup and dataset).

\smallskip
\textbf{Subgroup set discovery}. Subgroup set discovery\cite{van2012diverse} is the task of finding a set of high-quality, non-redundant subgroups that together describe all substantial deviations in the target distribution. That is, given 
a quality function $Q$ for subgroup sets and the set of all possible subgroup sets $\mathcal{S}$, the task is to find that subgroup set $S^* = \{s_1, \ldots ,s_k \}$ given by
$S^* = \argmax_{S \in \mathcal{S}} Q(S)$. 

Ideally this measure should 1) \emph{be global}, i.e., for a given dataset it should be possible to compare subgroup set qualities regardless of subgroup set size or coverage; 2) \emph{maximize the individual qualities} of the subgroups; and 3) \emph{minimize redundancy} of the subgroup set, i.e., the subgroups covers should overlap as little as possible while ensuring 2.

\section{MDL-based Subgroup Set Discovery}
\label{sec:MDL}

In this section we formalize the task of subgroup set discovery as a model selection problem using the Minimum Description Length (MDL) principle \cite{rissanen78,grunwald2007minimum}. To this end we first need to define an appropriate model class $\mathcal{M}$; as we will explain next, we use \emph{subgroup lists} as our models. The model selection problem should then be formalized using a two-part code \cite{grunwald2007minimum}, i.e., 
\begin{equation}\label{eq:LengthTotal}
M^* = \argmin_{M \in \M} L(D,M) = \argmin_{M \in \M} \left[ L(Y \given X,M)  + L(M) \right],
\end{equation}
where $L(Y \given X,M)$ is the encoded length, in bits\footnote{To obtain code lengths in bits, all logarithms in this paper are to the base 2.}, of target $Y$ given explanatory data $X$ and model $M$, $L(M)$ is the encoded length, in bits, of the model, and $L(D,M)$ is the total encoded length and the sum of both terms. Intuitively, the best model $M^*$ is that model that results in the best trade-off between how well the model compresses the target data and the complexity of that model---thus minimizing redundancy and automatically selecting the best subgroup list size. This formulation is similar to those previously used for two-view association discovery and multi-class classification \cite{van2015association,proencca2020interpretable}. We will first describe the details of the model class and then the required length functions.

\subsection{Model Class: Subgroup Lists}

Although Equation~\eqref{eq:LengthTotal} provides a \emph{global} criterion that enables the comparison of subgroup sets of different sizes, subgroups are descriptions of \emph{local} phenomena and we require each \emph{individual subgroup to have high quality}. 

We can accomplish this by using \emph{subgroup lists} as models; see Eq.~\eqref{eq:modelclass}. Specifically, as we are only interested in finding subgroups for which the target deviates from the overall distribution, we assume $y$ values to be distributed according to $\hat{f}_{d}$ by default (last line in Eq.~\eqref{eq:modelclass}). For each region in the data for which the target distribution deviates from that distribution and a description exists, a subgroup specifying a different distribution $\hat{f}_{a}$ is added to the list. 

We model the empirical distributions $\hat{f}$ by normal distributions, as those capture the two properties of interest, i.e., centre and spread, while being robust to cases where $f$ violates the normality assumption \cite{grunwald2007minimum}. We thus define $\hat{f}_{\hat{\mu},\hat{\sigma}}(y)= (2 \pi \hat{\sigma})^{-1/2} \exp{ \frac{(y-\hat{\mu})^2}{2 \hat{\sigma}^2}}$, where $\hat{\mu}$ and $\hat{\sigma}$ are the estimated mean and standard deviation, respectively. These statistics can be easily estimated using the maximum likelihood estimator, so that a pattern $a$ establishes a rule of the form $\textsc{if }   a \sqsubseteq  \x   \textsc{ then }  \mathcal{N}(\hmua,\hsiga)$. 
Combining subgroup distributions $\hat{f}_{a,\hat{\mu}_a,\hat{\sigma}_a}$ with estimated dataset distribution $\hat{f}_{d,\hat{\mu}_d,\hat{\sigma}_d}$, this leads to a subgroup list $M$ given by
\begin{equation}\label{eq:modelclass} \small
\begin{split}
\text{subgroup 1}: & \textsc{ if }   a_1 \sqsubseteq  \x   \textsc{ then }  \hat{f}_{a_1,\hat{\mu}_1,\hat{\sigma}_1}(y)\\ 
\vdots & \\ 
\text{subgroup k}:  &  \textsc{ else if }   a_k \sqsubseteq  \x   \textsc{ then }  \hat{f}_{a_k,\hat{\mu}_k,\hat{\sigma}_k}(y) \\  
\text{dataset}: &\textsc{ else }      \hat{f}_{d,\hat{\mu}_d,\hat{\sigma}_d}(y)
\end{split}
\end{equation}

This corresponds to a probabilistic rule list with $k=|S|$ subgroups and a last (default) rule which is fixed to the overall empirical distribution $\hat{f}_{d,\hat{\mu},\hat{\sigma}}$ \cite{proencca2020interpretable}. Fixing the distribution of this last `rule' is crucial and differentiates a subgroup list from rule lists as used in classification and/or regression, as this enforces the discovery of a set of subgroups that individually all have target distributions that substantially deviate from the overall target distribution.

\subsection{Model Encoding}

The next step is to define the two length functions; we start with $L(M)$. Following the MDL principle \cite{grunwald2007minimum}, we need to ensure that 1) all models in the model class, i.e., all subgroup lists for a given dataset, can be distinguished; and 2) larger code lengths are assigned to more complex models. To accomplish the former we encode all elements of a model that can change, while for the latter we resort to two different codes: when a larger value represents a larger complexity we use the universal code for integers \cite{grunwald2007minimum}, denoted\footnote{$\LN(i)= \log k_0 + \log^{\ast} i $, where $\log^{\ast} i = \log i + \log \log i + \ldots$ and $ k_0 \approx 2.865064$.} $\LN$, and when we have no prior knowledge but need to encode an element from a set we choose the uniform code.

Specifically, the encoded length of a model $M$ over variables $V$ is given by 
\begin{equation} \label{eq:LModel}
L(M) = L_\mathbb{N}(|S|) + \sum_{a_i \in S} \left[L_\mathbb{N}(|a_i|) +\log \binom{|V|}{|a_i|} + \sum_{v \in a_i} L(v)\right] ,
\end{equation}
where we first encode the number of subgroups $|S|$ using the universal code for integers, and then encode each subgroup description individually. For each description, first the number $|a_i|$ of variables used is encoded, then the set of variables using a uniform code over the set of all possible combinations of $|a_i|$ from $|V|$ variables, and finally the specific condition for a given variable. As we allow variables of three types, the latter is further specified by
\begin{equation}
L(v_{bin}) =\log 2 \: ; \: L(v_{nom}) =\log |\mathcal{X}_v| \:  ; \:  L(v_{num}) = \log N(n_{cut}),
\end{equation}

where the code for each variable type assigns code lengths proportional to the number of possible partitions of the variable's domain. Note that this seems justified, as more partitions implies more potential spurious associations with the target that we would like to avoid. For \texttt{bin}ary variables only two conditions are possible, while for \texttt{nom}inal variables this is given by the size of the domain. For \texttt{num}eric variables it equals the number of possible combinations $N(n_{cut})$, as there can be conditions with one (e.g. $ x \leq 2$) or two operators (e.g. $1 \leq x \leq 2$), which is a function of the number of possible subsets generated by $n_{cut}$ cut points. Note that we here assume that equal frequency binning is used, which means that knowing $X$ and $n_{cut}$ is sufficient to determine the cut points.

\subsection{Data encoding}

The remaining length function is that of the target data given the explanatory data and model,  $L(Y \given X,M)$. For this we first observe that for any given subgroup list of the form of Equation~\eqref{eq:modelclass}, \emph{any individual instance $(\x_i,y_i)$ is `covered' by only one subgroup}. That is, the cover of a subgroup $a_i$, denoted $\Da$, depends on the order of the list and is given by the instances where its description occurs minus those instances covered by previous subgroups:
\begin{equation}\label{eq:coversubgroup}
\Da =\{\Xa,\Ya  \} = \{(\x,y) \in D \given a_i \sqsubseteq \x \wedge \left ( \bigwedge_{\forall_{j<i}}  a_j \not \sqsubseteq \vec{x}   \right ) \}.
\end{equation}

Next, let $n_i =|\Da|$ be the number of instances covered by a subgroup (also known as \emph{usage}). For a given subgroup $a_i$, we then estimate 
\begin{multicols}{2}
  \begin{equation}\label{eq:mean}
    \hmua= \frac{1}{n_i}\sum_{y \in Y_i} y
  \end{equation}
  
  \begin{equation}\label{eq:std}
    \hsiga^2= \frac{1}{n_i}\sum_{y \in Y_i} (y-\hat{\mu}_i)^2,
  \end{equation}
\end{multicols}
where $\hsiga^2$ is the biased estimator such that the estimate times $n_i$ equals the Residual Sum of Squares, i.e., $n_i \hsiga^2 = \sum_{y \in \Ya} (y-\hmua)^2 = RSS_a $.

Given the above, we can separately encode the covers of the individual subgroups, but we first show how to encode the target values not covered by any subgroup.

\smallskip
\noindent \textbf{Encoding target values not covered by any subgroup.} The target values not covered by any subgroup, given by $Y_d = \{(\x,y) \in D \given \forall_{a_i \in M} a_i \not \sqsubseteq \vec{x}  \} $, are covered by the default dataset `rule' and distribution at the end of a subgroup list. As $\hat{f}_{d,\hat{\mu}_d,\hat{\sigma}_d}$ is known and constant for a given dataset, one can simply encode the instances using this (normal) distribution, resulting in  encoded length
\begin{equation}\label{eq:lengthdefault}
L(Y_d \given \hat{\mu}_d, \hat{\sigma}_d) = \frac{n_d}{2} \log 2\pi + \frac{n_d}{2} \log \hat{\sigma}_d^2 +  \left[ \frac{1}{2 \hat{\sigma}_d^2} \sum_{y \in Y_d} (y-\hat{\mu}_d)^2   \right] \loge,
\end{equation}

where $\loge = \log e$. The first two terms are normalizing terms of a normal distribution, while the last term represents the Residual Sum of Squares (RSS) normalized by the variance of the data. Note that when $Y_d = Y$, i.e., the whole dataset target, RSS is equal to $n_d \sigma_d $ and the last term reduces to $\loge n_d/2$.

\smallskip
\noindent \textbf{Encoding target values covered by a subgroup.} In contrast to the previous case, here \emph{we do not know a priori the statistics defining the probability distribution corresponding to the subgroup}, i.e., $\hat{\mu}$ and $\hat{\sigma}$ are not given by the model and thus both need to be encoded. For this we resort to the Bayesian encoding of a normal distribution with mean $\mu$ and standard deviation $\sigma$ unknown, which was shown to be asymptotically optimal \cite{grunwald2007minimum}. An optimal code length is simply given by the negative logarithm of a probability, and the optimal Bayesian probability for $\Ya$ is given by
\begin{equation}\label{eq:bayes}
 P_{Bayes} (\Ya) = \int_{-\infty}^{+\infty} \int_{0}^{+\infty} (2\pi \sigma)^{-\frac{n_i}{2}}\exp{- \frac{\sum_{y \in \Ya} (y - \mu)^2}{2 \sigma^2}}  w(\mu,\sigma) \dif \mu \dif \sigma,
\end{equation}
where $w(\mu,\sigma)$ is the prior on the parameters, which needs to be chosen.

The MDL principle requires the encoding to be as unbiased as possible for any values of the parameters, which leads to the use of uninformative priors. The most uninformative prior is  Jeffrey's prior, which is $1/\sigma^2$ and therefore constant for any value of $\mu$ and $\sigma$, but unfortunately its integral is undefined, i.e., $ \int \int \sigma^{-2} \dif \sigma \dif \mu = \infty$. Thus, we need to 1) constrain the parameter space and 2) make the integral finite, which we will do next in consecutive steps.

One of the best ways to constrain the parameter space without biasing it, is by multiplying  Jeffrey's prior by a normal prior on the effect size, i.e., $\rho = \mu/\sigma \sim \mathcal{N} (0,\tau)$ \cite{rouder2009bayesian}. We then still need to describe $\tau$ though; the most uninformative choice would be to use an inverse-chi-squared distribution, which would be equivalent to using a Cauchy prior on the effect size \cite{rouder2009bayesian}. Unfortunately, this would lead to an open integral, which would render the approach infeasible for cases---like ours---where many probabilities need to be computed. The second best option is to fix $\tau = 1$, which gives a tractable formula that is equivalent to introducing a virtual point and converges\footnote{See proof in Appendix~\ref{appendix:BIC}.} to the Bayes Information Criterion (BIC) for large $n$. This is the best we can do and we proceed with this option.

Now, given the prior defined by $\rho = \mu/\sigma \sim \mathcal{N} (0,1)$, the remaining question is how we can make the integral over the prior finite. The most common solution, which we also employ, is to use $k$ data points from $\Ya$, denoted $Y^k_i$, to create a proper conditional prior $w(\mu,\sigma \given Y^k_i)$. As there are only two unknown parameters, we only need two points hence $k=2$ \cite{grunwald2007minimum}, for more on the interpretation of such "priors conditional on initial data points", see \cite{grunwald2019minimum}. Consequently, we first encode $Y^2_i$ with a non-optimal code that is readily available---here the encoding with the dataset distribution of Equation~\eqref{eq:lengthdefault}---and then use the Bayesian rule to derive the total encoded length of $\Ya$ as
\begin{equation}
L(\Ya) = -\log \frac{P_{Bayes}(\Ya)}{P_{Bayes}(\Ytwo)}P(\Ytwo \given \mu_d,\sigma_d) = L_{Bayes}(\Ya)+L_{cost}(\Ytwo),
\end{equation} 
where $L_{cost}(\Ytwo) = L(\Ytwo \given \mu_d,\sigma_d) - L_{Bayes}(\Ytwo)$ is the extra cost incurred by encoding two points non-optimally. After some re-writing\footnote{The full derivation of the Bayesian encoding and an in-depth explanation are given in Appendix~\ref{appendix:Bayesian}.} we obtain the encoded length of the $y$ values covered by a subgroup $\Ya$ as
\begin{equation}\label{eq:lengthsubgroup}
\begin{split}
&L(\Ya)= L_{Bayes}(\Ya) + L_{cost}(\Ytwo)\\ 
&=1 +\frac{n_i}{2} \log \pi- \log \Gamma \left( \frac{n_i}{2} \right) + \frac{1}{2} \log (n_i+1) +\frac{n_i}{2} \log  n \hat{\sigma}_{a}^2  + L_{cost}(\Ytwo), \\
 \end{split}
\end{equation}
where $\Gamma$ is the Gamma function that extends the factorial to the real numbers ($\Gamma (n) = (n-1)!$ for integer $n$) and $\hmua$ and $\hsiga$ are the statistics of Equations~\eqref{eq:mean} and \eqref{eq:std}, respectively. Note that for $\Ytwo$ any two unequal values (otherwise $\hat{\sigma}_2 =0$ and $L_{Bayes}(\Ytwo) = \infty$) can be chosen from $\Ya$, thus we choose them such that they minimize $L_{cost}(\Ytwo)$.  

Finally, the total encoded size of $Y$ is given by 
\begin{equation}
L(Y \given X, M) = \sum_{i \in M}  L(\Ya) + L(Y_d \given \mu_d, \sigma_d).
\end{equation}

\subsection{Properties and quality measure for subgroup lists}\label{section:MDLsubgroup}

We next show\footnote{Derivations are given in Appendix~\ref{appendix:WKL}.} that the proposed data encoding is an instance of the classical definition of a quality measure as given by Equation~\eqref{eq:generalmeasure}, and is tightly related to both an existing quality measure and the Bayesian two-sample t-test.

First, we show that Equation~\eqref{eq:lengthsubgroup}---with mean and variance unknown---converges, for large $n$, to Equation~\ref{eq:lengthdefault}---with mean and variance known---plus an additional term. Using the Stirling approximation of $\Gamma(n+1) \sim  \sqrt{2\pi n } \left( \frac{n}{e} \right)^n$ leads to
\begin{equation}
 L (\Ya) \sim \frac{n_i}{2} \log 2\pi + \frac{n_i}{2} \log \hsiga^2 +  \frac{n_i}{2} \loge + \log \frac{n_i}{e},
\end{equation}
where $\log \frac{n}{e}$ is equal to the penalty term of BIC and similar to the usual MDL complexity of a distribution \cite{grunwald2007minimum}.

Now, we can show that minimizing our MDL criterion is equivalent to maximizing a subgroup discovery quality function of the form Eq~\eqref{eq:generalmeasure}. Focusing on the case where $S = \{s_1\}$ contains only one subgroup with statistics $ \hat{\Theta}_1 =\{\hat{\mu}_1,\hat{\sigma}_1 \}$, we start with $L(Y \given X,M)$ (Eq.~\eqref{eq:LengthTotal}), multiply it by minus one to make it a maximization problem, and add a constant $L(Y \given  \hat{\mu}_d, \hat{\sigma}_d)$, i.e., the encoded size of the whole target $Y$ using the overall distribution dataset, to obtain
\begin{equation}\label{eq:KLproof} \small
\begin{split}
L(Y \given  \hat{\Theta}_d)-L(Y \given X,M) &\sim  
n_i \left [ \log\frac{\hat{\sigma}_d}{\hsiga}+ \frac{\hsiga^2+(\mu_1-\mu_2)^2}{2 \sigma_d^2}\loge -\frac{\loge}{2} \right ] - \log(n_i) - L(S)  \\
&=n_i D_{KL}(\hat{\Theta}_a;\hat{\Theta}_d) - \log(n_i)- L(S),
\end{split}
\end{equation}

where $\hat{\Theta}_a = \{\hat{\mu}_d, \hat{\sigma}_d\}$ and $n_i D_{KL}(\hat{\Theta}_a; \hat{\Theta}_d)$ is the usage-weighted Kullback-Leibler divergence between the normal distributions specified by the respective parameter vectors \footnote{As shown in Appendix~\ref{appendix:KLD}}. This shows that \emph{finding the MDL-optimal subgroup is equivalent to finding the subgroup that maximizes the weighted Kullback-Leibler (WKL) divergence}, an existing subgroup discovery quality measure~\cite{van2010maximal} that was previously used for nominal targets, plus a term that defines the complexity of the subgroup. Moreover, note that Eq.~\eqref{eq:KLproof} is equivalent to the Bayesian two-sample t-test \cite{gonen2005bayesian} plus the complexity of the model, which plays the role of penalizing for multiple hypothesis testing. Finally, our measure is part of the family of \emph{dispersion-corrected} subgroup quality measures, as it takes into account both the centrality and the spread of the target values \cite{boley2017identifying}.

\smallskip
\noindent \textbf{Quality measure for subgroup lists}. Based on the previous, we naturally extend the KL-based measure for individual subgroups to subgroup lists and propose the Sum of Weighted Kullback-Leibler (SWKL) divergences:
\begin{equation}\label{eq:swkl}
\textrm{SWKL} (S) = \sum_{a\in S} n_i D_{KL}(\hat{\Theta}_a;\hat{\Theta}_d) =  \sum_{a_i\in S} n_i \left [ \log\frac{\hsigd}{\hsiga}+ \frac{\hsiga^2+(\hmua-\hmud)^2}{2 \hsigd^2}\loge -\frac{\loge}{2} \right ]
\end{equation}

An advantage of this measure is that it can not only be used for numeric targets, but for any type of probabilistic model. Note that computing SWKL is straightforward for subgroup lists as obtained by most methods, including ours, but not for subgroup sets as instances can be covered by multiple subgroups. In those cases, it is necessary to explicitly define the type of probabilistic overlap, e.g., additive or multiplicative mixtures of the individual subgroup models.
\section{The SSD++ Algorithm}

As the problem of finding an MDL-optimal list of subgroups is unfeasible, we propose a heuristic approach  (as is common in MDL-based pattern mining \cite{vreeken2011krimp,proencca2020interpretable}) based on Separate-and-Conquer (SaC) to construct the list, and beam-search to generate the subgroups to add at each iteration of SaC. The first reason for using greedy search to add one subgroup at the time, is its transparency, as it adds at each iteration the locally best subgroup found by the beam search. Beam-search, on the other hand, was empirically shown, in the context of subgroup discovery for numeric targets, to be very competitive in terms of quality when compared to a complete search with an associated speedup improvement \cite{meeng2020forreal}. Also, its straightforward implementation allows to easily extend this framework to other types of targets, not just numeric. To quantify the quality of annexing $\oplus$ a subgroup $s$ at the end (after all the other subgroups) of model $M$, we employ the \emph{normalized gain} $\delta L(M \oplus s) = (L(D,M) - L(D,M\oplus s))/n_s$, which was first introduced in the classification setting and proved to perform better than its non-normalized version in that setting \cite{proencca2020interpretable}. For a detailed empirical comparison of normalized gain and its non-normalized version please refer to Appendix~\ref{appendix:empiricalabsvsnorm}. 

Note that this gain is a normalized version of equation~\eqref{eq:KLproof}, and as such, all subgroups selected in this way are maximizing a normalized version of a Bayesian two-sample t-test (plus a multiple hypothesis penalization), hence they are all individually ``significant'' according to this test.\\

Algorithm~\ref{alg:SSD} presents SSD++, a greedy algorithm that starts with an empty subgroup list and iteratively adds subgroups until no more compression can be gained, where compression is measured in terms of normalized gain of adding a subgroup $s$. 

The \emph{beam search algorithm} starts by discretizing all variables depending on their subsets, i.e. categorical and binary with the operator \emph{equal to} ($=$) and numeric by generating all subsets with $n_{cut}$ points. At  each iteration the $w_b$ subgroups that maximize the selected gain are chosen and will be expanded with all discretized variables until the maximum depth $d_{max}$ of the description is achieved. 

The \emph{SSD++ algorithm}\footnote{Code is publicly available here: \url{https://github.com/HMProenca/SSDpp-numeric}} starts by taking as input the dataset $D$, and the beam search parameters, namely the number of cut points $n_{cut}$, the width of the beam $w_b$, and the maximum depth of search $d_{max}$. The algorithm starts by adding the dataset empirical distribution to the model (Ln~\ref{alg:initialize}). Then, while there is a subgroup that improves compression (Ln~\ref{alg:loop}), it keeps iterating over three steps: 1) generating the candidates using beam search (Ln~\ref{alg:beam}); 2) finding the subgroup that maximizes the normalized gain (Ln~\ref{alg:gain}); and 3) adding that subgroup to the  end of the model, i.e., after all the existing subgroups in the model (Ln~\ref{alg:addrule}). The beam search returns the best subgroup according to the data not covered by any subgroup in the model $M$ and its parameters ($w_b,n_{cut},d_{max})$. When there is no subgroup that improves compression (non-positive gain) the while loop stops and the subgroup list is returned. Note that beam search is used at each iteration, instead of only once at the beginning, as it can converge to local optima, and would thus bias our search to the top-k subgroups instead of the best at each iteration. 
\begin{algorithm}[!t]
    \centering
	\caption{SSD++ algorithm}\label{alg:SSD}
	\begin{algorithmic}[1]
		\INPUT Dataset $D$, number of cut points $n_{cut}$, beam width $w_b$, depth max. $d_{max}$
		\OUTPUT Subgroup list $S$
		\State $M \gets [\Theta_d(Y)]$ \label{alg:initialize}
		\Repeat \label{alg:while}
		\State $\Cands \gets BeamSearch(M,D,w_b,n_{cut},d_{max})$ \label{alg:beam}
		\State $s \gets \argmax_{\forall s' \in \Cands} :\dn (D,M \oplus s')$ \label{alg:gain}		
		\State $M  \gets M \oplus s$	\label{alg:addrule}	
		\Until{$\dn (D,M \oplus s') \leq 0, \forall s' \in \Cands$}\label{alg:loop}
		\State \textbf{return} $S \in M$
	\end{algorithmic}
\end{algorithm} 

\section{Experiments}
\label{section:experiments}

We evaluate SSD++ by comparing it to 1) a classical top-k mining algorithm, as a baseline of a non-diverse method, and 2) the sequential covering algorithm, henceforth called top-k and seq-cover respectively, which are both available in the implementation of the the DSSD algorithm\footnote{\url{http://www.patternsthatmatter.org/software.php\#dssd/}}.

DSSD and SISD  will not be compared due to two interconnected issues: 1) the lack of a \emph{global} definition of the optimal set for a dataset; 2) the absence of a definition for the interaction between subgroups that overlap. The first issue has as a natural consequence that none of the methods have a clear stopping criteria as the definition of when a set describes the data well is not available, apart from the user-specified hyperparameter `number of subgroups'. Added to this, both issues give rise to the question of how to measure the interaction of subgroups in the region of their overlap from a model (global) perspective, i.e., they could behave as an additive or a multiplicative mixture of their probabilities for example. These issues hamper the comparison with both methods as they do not have a clear stopping criteria and a formulation of their overlap interaction, of which the latter is necessary for our proposed measure SWKL. On the other hand, a direct use of SWKL assuming a list formulation, i.e. ordering them and removing the overlap, will always rate them lower, which was corroborated with our initial experiments.
Note that we also do not compare with machine learning algorithms that generate rules for regression, such as RIPPER or CART, as the rules generated aim at making the best prediction possible, and not the highest difference from the dataset distribution, as shown theoretically in Appendix~\ref{appendix:proof_sdvsregression}.

\textbf{Data} We use a set of $16$ benchmark datasets from the Keel\footnote{\url{http://www.keel.es/}} repository commonly used for subgroup discovery. The complete description of the datasets is given in Table~\ref{table:datasets}; the datasets were chosen to be diverse, ranging from $297$  to $22 \; 784$ instances and from $2$ to $40$ variables. It should be noted that most datasets do not have categorical variables as this is not common in numeric/regression settings. 

\textbf{Hyperparameter selection.} \emph{SSD++:} the algorithm admits as hyperparameters: the width of the beam $w_b$; number of cut points $n_{cut}$; and maximum depth of search $d_{max}$. 
By varying these parameters over the datasets the results can be seen in Appendix~\ref{appendix:empiricalbeamsearch} and it was concluded that: 1) no descriptions of size much greater than $5$ are found; 2) after $n_{cut} = 5$ (the default value for seq-cover) the subgroups returned are virtually the same but with numerical values refined; 3) for most datasets the quality of the subgroup list stabilizes beyond $w_b=100$. Thus, for the rest of our experiments the parameters are set according to these findings.

\begin{table}[!t]\centering  		\caption{Dataset properties: number of instances, and variables.}\label{table:datasets} 																	
	\ra{1.1} \begin{tabular}{@{}lrrrclrrr@{}}\toprule																		
		Dataset	&	$|D|$	&	categorical	&	numerical	&	&	Dataset	&	$|D|$	&	categorical	&	numerical		\\  \midrule
		cholesterol	&$	297	$&$	7	$&$	5	$&\phantom{a}	&	wizmir 	&$	1\;461	$&$	0	$&$	9	$	\\  
		baseball 	&$	337	$&$	4	$&$	12	$&	&	abalone 	&$	4\,177	$&$	0	$&$	8	$	\\ 
		autoMPG8 	&$	392	$&$	0	$&$	6	$&	&	puma32h 	&$	8\,192	$&$	0	$&$	32	$	\\ 
		dee 	&$	365	$&$	0	$&$	6	$&	&	ailerons 	&$	13\,750	$&$	0	$&$	40	$	\\ 
		ele-1 	&$	495	$&$	0	$&$	2	$&	&	elevators 	&$	16\,599	$&$	0	$&$	18	$	\\ 
		forestFires 	&$	517	$&$	0	$&$	12	$&	&	bikesharing	&$	17\,379	$&$	2	$&$	10	$	\\ 
		concrete 	&$	1\,030	$&$	0	$&$	8	$&	&	california 	&$	20\,640	$&$	0	$&$	8	$	\\ 
		treasury 	&$	1\,049	$&$	0	$&$	15	$&	&	house 	&$	22\,784	$&$	0	$&$	16	$	\\ 
		\bottomrule																	
	\end{tabular}
	
\end{table}

\emph{Top-k:} the software used here is the top-k subgroups implemented in DSSD, which is equivalent to most top-k subgroup miners. As it is common with top-k miners a depth-first search is used for small datasets $|D| \leq 2000$ and a beam search for the rest. For the quality measure it uses the Weighted Kullback-Leibler without dispersion, i.e., $WKL_{\mu}(s) = n_s/\hat{\sigma}_d (\hat{\mu_d}-\hat{\mu_s})^2$ as described in Appendix 3, as the algorithm does not accept its dispersion-aware version used in Eq.~\eqref{eq:swkl}. Also, as it does not have a termination criteria, the $k$ number of subgroups returned is selected as the number of subgroups found by SSD++.

\emph{Seq-cover:} to ensure fairness the same beam search hyperarameters as SSD++ are used, i.e., $d_{max} = 5$, $w_b = 100$, $n_{cut} =5$. As quality measure it uses the Weighted Kullback-Leibler without dispersion for the same reasons as top-k. Even though some versions of sequential covering use a form of exhaustive search at each iteration, the use of beam search instead should not quantitatively deteriorate the results as shown by Meeng and Knobbe~\cite{meeng2020forreal} when comparing both search methods in subgroup discovery.

\subsection{Subgroup List Quality}

\begin{figure}[h!]
	\begin{minipage}[t]{0.48\textwidth}
		\includegraphics[width=1.0\textwidth]{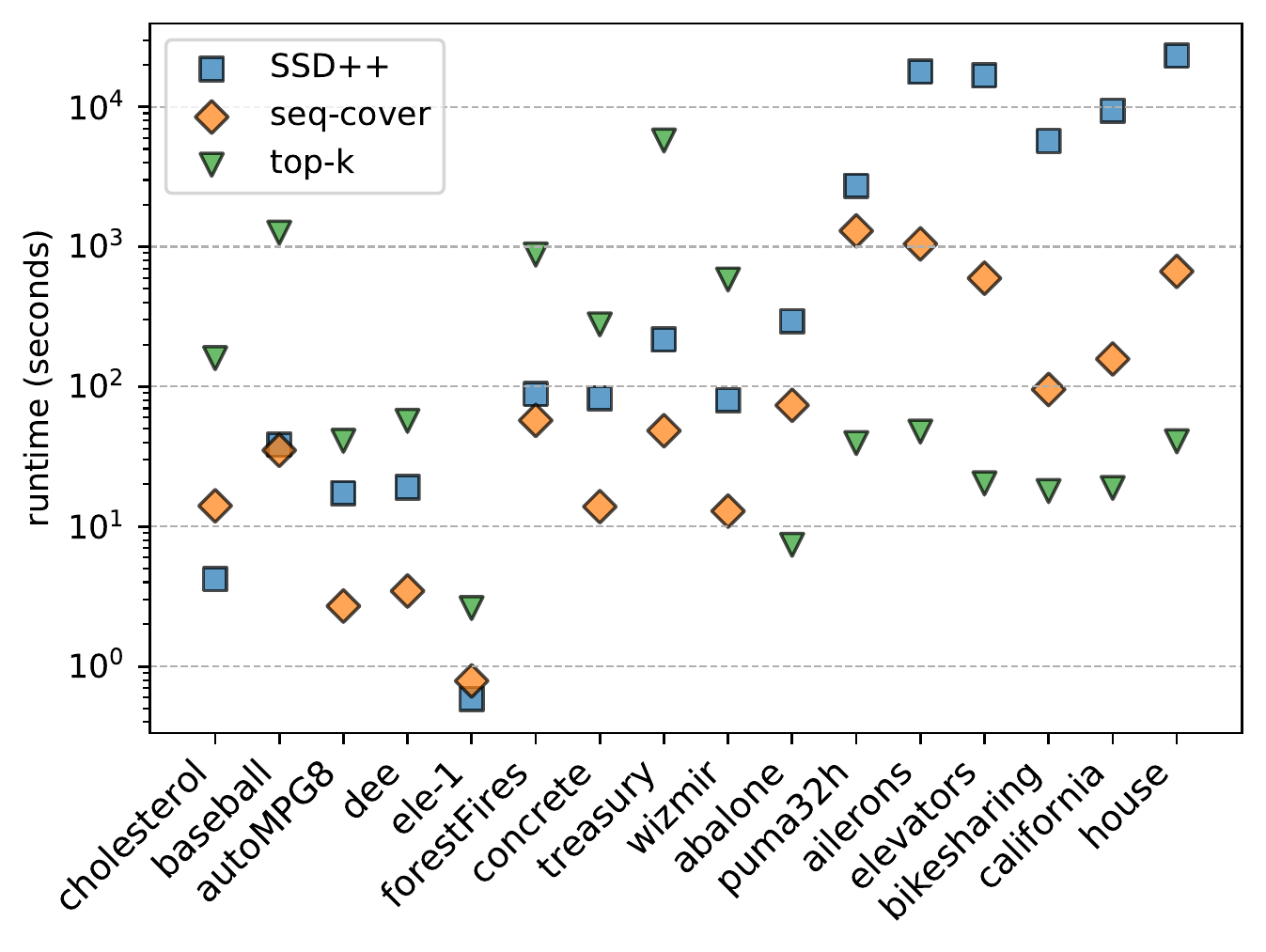} 
		\caption{Runtime in seconds for all the algorithms for each dataset.}\label{fig:runtime}
	\end{minipage}\hfill
	\begin{minipage}[t]{0.48\textwidth}
		\includegraphics[width=1.0\textwidth]{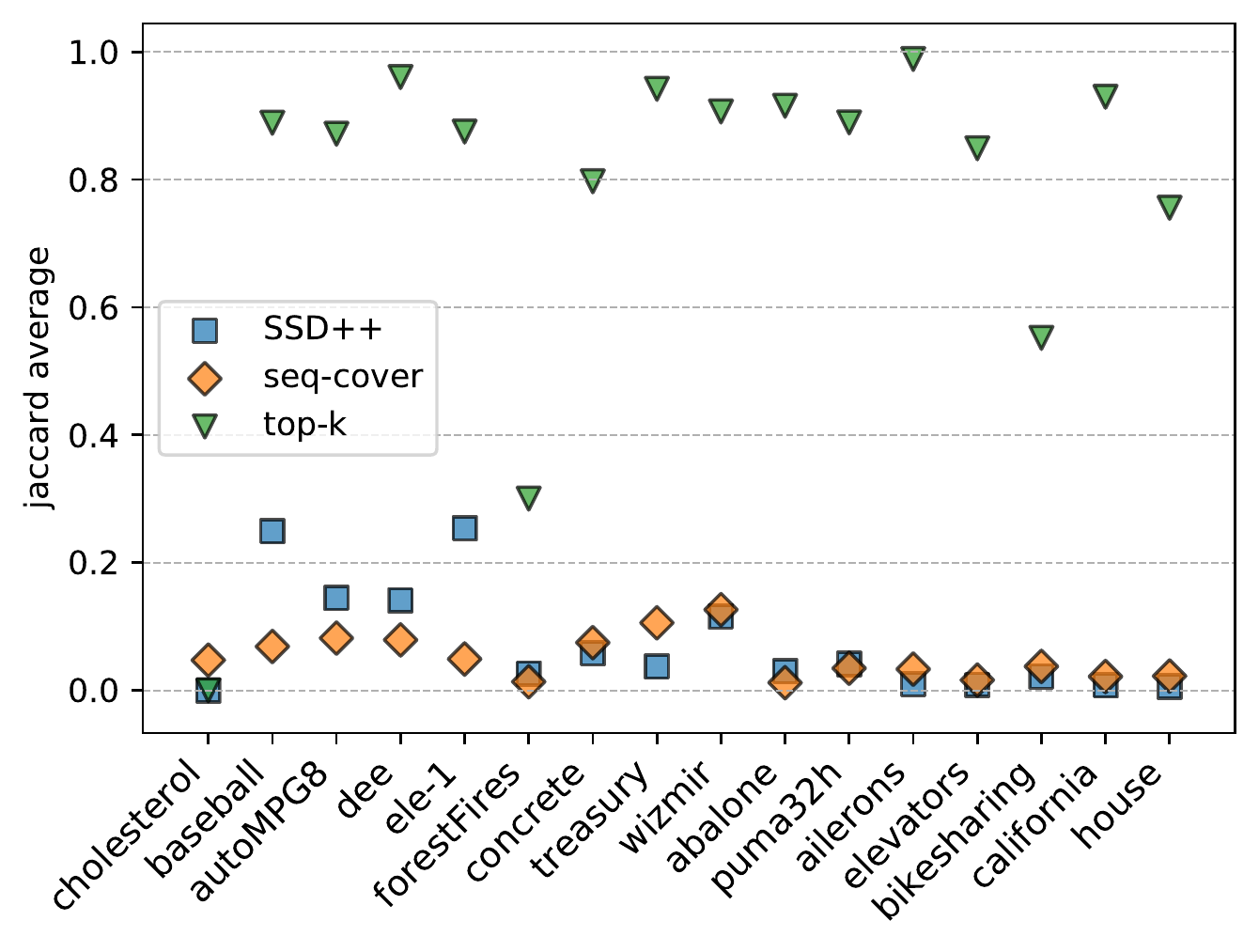} 
		\caption{Average overlap over each two subgroups (Jaccard index average) for each dataset and algorithm.}\label{fig:jaccard}
	\end{minipage}
\end{figure}

The results can be seen in Table~\ref{table:swkl}, and Figures~\ref{fig:runtime} and \ref{fig:jaccard}. The algorithms are compared in terms of Sum of Weighted Kullback-Leibler (SWKL) of Eq.~\ref{eq:swkl} for the quality of the list, number of subgroups $|S|$, average number of variables per description $|a|$, standard deviation of the first subgroup $\Tilde{\sigma}_{top1}$, runtime and average Jaccard index of the lists. Note that $\Tilde{\sigma}_{top1}$ is used as it shows what is the most important characteristic first found by each miner. In the case of the averaged Jaccard index it is computed based on the average of the Jaccard index between the 1-vs-1 covers (when considered independently) of the subgroups in the list, i.e., for the case of a list of $4$ subgroups, $6$ values are averaged.

From Table~\ref{table:swkl} we can see that SSD++ obtains the best score in terms of our proposed measure SWKL for $12$ out of $16$ datasets. As expected the top-k algorithm obtains a lower score for all datasets except for one. This  supports that our proposed measure SWKL gives weight to subgroup sets that cover different parts of the dataset. Also, in terms of the dispersion of the first subgroup its value is lower for $80\%$ of the cases.
In terms of the number of rules and compared with seq-cover, SSD++ tends to find fewer subgroups for smaller datasets ($|D| \leq 10\: 000$), and more for larger datasets. For the latter, the experiments showed that on average each subgroup covers more than $100$ instances per subgroup. In terms of the number of variables per description, it tends to find more compact descriptions than top-k and seq-cover.

In terms of runtime, as per Figure~\ref{fig:runtime}, SSD++ has a similar performance to seq-cover for small sample sizes ($|D| \leq 1000$) and $10$ times slower for larger sizes. This can, in part, be explained, by the larger number of subgroups found for these datasets---from $1.2$ to $4$ times more. Figure~\ref{fig:jaccard} shows that for small datasets the overlap is larger than for seq-cover, while for larger datasets our formulation tends to have similar level of overlap.
\begin{table*}[t!]\centering					
\caption{Performance results of \{Summed Weighted Kullback-Leibler Divergence (SWKL) divided by number of examples; standard deviation of the first subgroup normalized by $\sigma_d$; number of subgroups; average number of conditions per subgroup description\} per dataset for each algorithm.}\label{table:swkl}		
\ra{1.1} \begin{tabular}{@{}lrrrrrrrrrrrrrr@{}}\toprule
	&	\multicolumn{4}{r}{top-k}							&	&	\multicolumn{4}{r}{seq-cover}							&	&	\multicolumn{4}{r}{SSD++}							\\	\cmidrule(l){2-5} \cmidrule(l){7-10}\cmidrule(l){12-15}
datasets	&\small	SWKL	&	$\tilde{\sigma}_{top1}$	&	$|S|$	&\phantom{a}	$|a|$	&	&\small	SWKL	&	$\tilde{\sigma}_{top1}$	&	$|S|$	&\phantom{a}	$|a|$	&	&\small	SWKL	&	$\tilde{\sigma}_{top1}$	&	$|S|$	&\phantom{a}	$|a|$	\\	\hline
cholesterol	&$	0.14	$&$\phantom{a}	1.49	$&$	1	$&$	5	$&	&$\pmb{	0.84	}$&$\pmb{	1.51	}$&$	33	$&$	4	$&	&$	0.11	$&$\phantom{a}	1.99	$&$	1	$&$	3	$\\	
baseball 	&$	0.25	$&$	0.85	$&$	8	$&$	5	$&	&$	1.69	$&$	0.82	$&$	26	$&$	4	$&	&$\pmb{	1.92	}$&$\pmb{	0.22	}$&$	8	$&$	2	$\\	
autoMPG8 	&$	0.48	$&$	0.54	$&$	10	$&$	5	$&	&$	1.36	$&$	0.54	$&$	22	$&$	3	$&	&$\pmb{	1.65	}$&$\pmb{	0.18	}$&$	10	$&$	2	$\\	
dee 	&$	0.49	$&$	0.47	$&$	8	$&$	5	$&	&$\pmb{	1.47	}$&$	0.50	$&$	20	$&$	4	$&	&$	1.33	$&$\pmb{	0.44	}$&$	8	$&$	2	$\\	
ele-1 	&$	0.29	$&$\pmb{	1.06	}$&$	9	$&$	3	$&	&$	1.14	$&$\pmb{	1.06	}$&$	22	$&$	3	$&	&$\pmb{	1.25	}$&$	1.33	$&$	9	$&$	2	$\\	
forestFires 	&$	0.58	$&$	6.84	$&$	23	$&$	5	$&	&$	2.85	$&$	6.84	$&$	57	$&$	4	$&	&$\pmb{	3.80	}$&$\pmb{	0.03	}$&$	23	$&$	3	$\\	
concrete 	&$	0.25	$&$	0.78	$&$	19	$&$	5	$&	&$\pmb{	1.27	}$&$	0.65	$&$	35	$&$	4	$&	&$\pmb{	1.27	}$&$\pmb{	0.34	}$&$	19	$&$	3	$\\	
treasury 	&$	0.42	$&$	0.70	$&$	31	$&$	5	$&	&$	2.41	$&$	0.68	$&$	25	$&$	3	$&	&$\pmb{	3.73	}$&$\pmb{	0.05	}$&$	31	$&$	2	$\\	
wizmir 	&$	0.77	$&$	0.31	$&$	22	$&$	5	$&	&$	2.17	$&$	0.31	$&$	26	$&$	4	$&	&$\pmb{	2.73	}$&$\pmb{	0.16	}$&$	22	$&$	2	$\\	
abalone 	&$	0.23	$&$	0.59	$&$	25	$&$	5	$&	&$	0.48	$&$	0.59	$&$	118	$&$	3	$&	&$\pmb{	0.71	}$&$\pmb{	0.45	}$&$	25	$&$	3	$\\	
puma32h 	&$	0.55	$&$	0.59	$&$	42	$&$	5	$&	&$\pmb{	1.48	}$&$	0.59	$&$	76	$&$	5	$&	&$	1.42	$&$\pmb{	0.30	}$&$	42	$&$	3	$\\	
ailerons 	&$	0.24	$&$	1.23	$&$	19	$&$	2	$&	&$	1.04	$&$	1.23	$&$\phantom{a}	101	$&$	4	$&	&$\pmb{	1.58	}$&$\pmb{	1.10	}$&$	197	$&$	4	$\\	
elevators 	&$	0.25	$&$	\pmb{1.44}	$&$	141	$&$	4	$&	&$	0.84	$&$\pmb{	1.44	}$&$	157	$&$	4	$&	&$\pmb{	1.30	}$&$\pmb{	1.44	}$&$	160	$&$	4	$\\	
bikesharing	&$	0.27	$&$	1.09	$&$	127	$&$	5	$&	&$	1.24	$&$\phantom{a}	1.09	$&$	91	$&$	4	$&	&$\pmb{	1.68	}$&$\pmb{	0.07	}$&$	127	$&$	4	$\\	
california 	&$	0.19	$&$	0.90	$&$	163	$&$	4	$&	&$	0.70	$&$	0.90	$&$	135	$&$	5	$&	&$\pmb{	1.15	}$&$\pmb{	0.84	}$&$	163	$&$	4	$\\	
house 	&$	0.19	$&$\pmb{	1.59	}$&$\phantom{a}	280	$&$	5	$&	&$	0.91	$&$\pmb{	1.59	}$&$	145	$&$	4	$&	&$\pmb{	2.08	}$&$	2.18	$&$\phantom{a}	280	$&$\phantom{a}	5	$\\	
\bottomrule	
\end{tabular}								
\end{table*}

\section{Case Study: Hotel Bookings}\label{section:casestudy}
\begin{figure}[h!]
	\centering
	\begin{minipage}{0.70\textwidth}
		\includegraphics[width=1.0\textwidth]{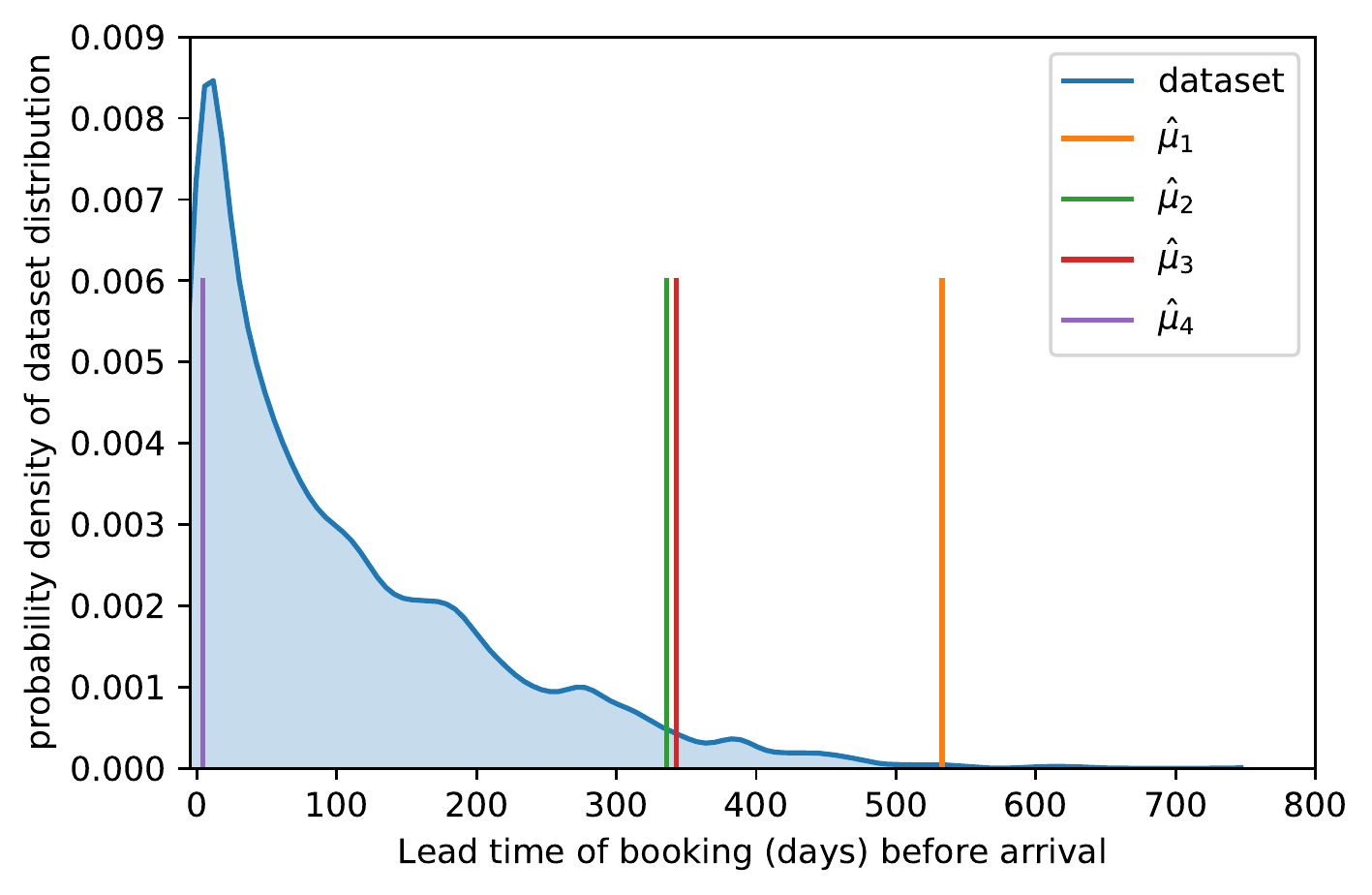} 
	\end{minipage}
\caption{Kernel density estimation of the dataset distribution and location (not density) of the mean value of the first $4$ subgroups.}\label{fig:hotelsubgroups}
\end{figure}

To test the usefulness of our method we applied it to the problem of understanding the type of clients that make a hotel booking based on how much time in advance (lead time in days) this was done. To this end we used the ``Hotel booking demand dataset''\cite{antonio2019hotel}, and analysed the data referent to a \emph{resort hotel} in the year of $2016$. The first four subgroups of a total of $260$ obtained with SSD++ can be seen in Figure~\ref{fig:hotelnorm_application} (in Section~\ref{section:intro}) and its subgroups versus the dataset in Figure~\ref{fig:hotelsubgroups}. Only the first $4$ subgroups are shown here for clarity, and given that greedy search is used, they are also the $4$ most interesting subgroups.

The results show us a very detailed picture of the dataset and at first glance, one notices that most subgroups cover a small number of instances. Nevertheless, this is normal as they represent highly defined subgroups, with a very different mean and an almost zero standard deviation, compared with the dataset $\hat{\mu}_d =92$ and $\hat{\sigma}_d = 99$. As an example, subgroup $1$ has an average lead time circa $6$ times higher than the dataset distribution, together with a standard deviation that is $3$ times smaller. This subgroup seems to represent a group of people that travelled together from Great Britain and all chose the same type of booking, while with some slight days of difference in their bookings. Another interesting subgroup is the $4^{th}$ which shows that there is a group of around $20$ similar bookings for groups of $2$ or more adults done with only $9$ days before arrival when the deposit type is refundable. If one would follow the whole subgroup list one would have a complete summary of the bookings done. 
\section{Conclusions}

We introduced a dispersion-aware problem formulation for subgroup set discovery based on subgroup lists, the MDL principle, and Bayesian statistics. We proved our  formulation to be equivalent to an existing subgroup quality measure for the case of finding the single best subgroup, and showed a relationship to Bayesian testing. Based on these insights we proposed a new evaluation measure for subgroup lists, the sum of Weighted Kullback-Leibler divergences (SWKL).

To find good subgroup lists we introduced SSD++, a greedy algorithm that we empirically evaluated on $16$ datasets and compared against state-of-the-art algorithms. SSD++ was shown to outperform the other methods in terms of both our proposed measure and subgroup set complexity as quantified by subgroup and/or description sizes, and discovers subgroups with small standard deviation. \\

\textbf{Future work} includes \textit{user defined constraints} to guide the search, such as minimum coverage for each subgroup and maximum number of subgroups in the list, and an \textit{extension of the MDL formulation for subgroup lists} to nominal and multiple targets.

\section*{Acknowledgments} This work is part of the research programme Indo-Dutch Joint Research Programme for ICT 2014 with project number 629.002.201, SAPPAO, which is (partly) financed by the Netherlands Organisation for Scientific Research (NWO).

\bibliographystyle{splncs04}
\bibliography{bibliography}
\begin{appendix}
\chapter*{Appendices}
\section{Bayesian encoding of a normal distribution with mean and standard deviation unknown}\label{appendix:Bayesian}

For encoding a sequence of numeric valued i.i.d. observations such as $Y_n = (y_1,....,y_n)$, the Bayesian encoding takes the following form:
\begin{equation}
P_{Bayes} (Y_n) = \int_{\Theta} f (Y \given \Theta) w(\Theta) \dif \Theta,
\end{equation}
where $f$ is the probability density function (pdf), $\Theta$ is the set of parameters of the distribution, and $w(\Theta)$ the prior over the parameters. 
In the case of a normal distribution $\Theta = \{\mu, \sigma \}$, with $\mu$ and $\sigma$ being its mean and standard deviation, respectively, the pdf $f(Y_n \given \Theta)$ over a sequence $Y_n$ is the multiplication of the individual pdfs, thus:
\begin{equation}
f(Y_n \given \mu, \sigma) = \frac{1}{(2 \pi)^{n/2} \sigma^n } \exp \left[ -\frac{1}{2 \sigma^2} \sum_i^n (y_i-\mu)^2   \right],
\end{equation}

In order not to bias the encoding for specific values of the parameters, we choose to use a normal prior on the effect size $\rho = \mu/\sigma$ and the constant Jeffrey's prior of $1/\sigma^2$ for the unknown parameters $\mu$ and $\sigma$. Thus, our prior is given by:
\begin{equation}
w(\mu,\sigma) = \frac{1}{\sqrt{2 \pi} \tau \sigma^2 } \exp \left[ -\frac{1}{2 \sigma^2} \frac{\mu^2}{\tau^2}  \right].
\end{equation}

Putting everything together, one obtains:
\begin{equation}\label{eq:BayesStep2}
\begin{split}
&P_{Bayes} (Y_n)= \\
&= (2 \pi)^{-\frac{n+1}{2}}\tau^{-1}  \int_{-\infty}^{+\infty}\int_{0}^{+\infty} \frac{1}{\sigma^{n+2} } \exp \left[ -\frac{1}{2 \sigma^2} \left(  \sum_i^n (y_i-\mu)^2 +\frac{\mu^2}{\tau^2} \right) \right] \dif \sigma \dif \mu.
\end{split}
\end{equation}

The integrals over the whole space of the parameters $\mu$ and $\sigma$ allow to penalize the fact that we do not know the statistics \emph{a priori}, thus penalizing the fact that a distribution over $n$ points could, by chance, have the same statistics as the one found in the data. 
Note that this prior choice is equal to the one of G\"{o}nen et al.\cite{gonen2005bayesian} for the Bayesian two-sample t-test, which was shown to converge to the Bayes Information Criteria (BIC) for large $n$ \cite{rouder2009bayesian}.

Note that using an improper prior requires that we somehow make it proper, i.e., we need to find a way to make the integration over the prior finite $ \int \int w(\mu, \sigma) = K,$ where $K$ is a constant value. The usual way to make an improper prior finite is to condition on the $k$ minimum number observations $Y^k_n \in Y_n$ needed to make the integral proper \cite{grunwald2007minimum}, which in the case of two unknowns ($\mu$ and $\sigma$) is $k =2$. Thus, instead of using $w(\mu,\sigma)$ we will in practice be using $w(\mu,\sigma \given \Ytwon)$, and using the the chain rule and the Bayesian formula returns a total encoding of $Y_n$ equal to
\begin{equation}\label{eq:conditionalbayes}
P(Y_n) = P_{Bayes} (Y_n \given \Ytwon)P (\Ytwon) = \frac{P_{Bayes} (Y_n)}{P_{Bayes} (\Ytwon)} P (\Ytwon)
\end{equation}
where $P (\Ytwon)$ is a non-optimal probability used to define $\Ytwon = \{ y_1,y_2 \}$ that we will define later and $y_1,y_2$ chosen in a way that maximizes $P(Y_n)$. 
Now that we have all the ingredients to define $P(Y_n)$ we will start by defining $P_{Bayes} (Y_n)$ and then choose the appropriate probability for $P(\Ytwon)$.

To solve the first integral of $P_{Bayes} (Y_n)$ in Equation~\eqref{eq:BayesStep2}, we integrate in $\sigma$ and note that the formula is an instance of the gamma function,
\begin{equation}
\Gamma (k) = \int_{0}^{+\infty} z^{k-1} e^{-z} \dif z,
\end{equation}
with the corresponding variable transformation:
\begin{equation}
z= \frac{A}{2 \sigma^2} ;\; \frac{1}{\sigma} = \frac{2^{1/2}z^{1/2}}{A^{1/2}} ;\;  \dif \sigma = - \frac{\sigma}{2z} \dif z ;\;  A = \left[ \sum_i^n (y_i-\mu)^2 +\frac{\mu^2}{\tau_\rho^2} \right],
\end{equation}
Performing the variable transformation and noting that the minus sign of $\dif z$ cancels with the reversing of the integral limits, we get:
\begin{equation}\label{eq:BayesStep3}
\begin{split}
&P_{Bayes} (Y)= \\
&= \tau^{-1} \Gamma(n/2) 2^{\frac{n+1}{2}-1} (2 \pi)^{-\frac{n+1}{2}} \int_{-\infty}^{+\infty} \left[ \sum_i^{n} (y_i-\mu)^2 +\frac{1}{\tau^2}(0-\mu)^2 \right]^{-\frac{n+1}{2}} \dif \mu ,
\end{split}
\end{equation}
which reveals that the prior on the effect size $\rho$, and specifically its standard deviation parameter $\tau$, is equivalent to adding $1/\tau^2$ virtual points to the original data.

To solve the integral in $\mu$ we need to introduce the statistics $\hat{\mu}$ and $\hat{\sigma}$ as the values estimated from the data. We define these quantities as:
\begin{equation}
\hat{\mu}= \frac{1}{n} \sum_i^n y_i ; \; \hat{\mu}'= \frac{n}{n+1/\tau^2} \hat{\mu}; \;  \hat{\sigma}^2 = \frac{1}{n} \sum_i^n (y_i-\hat{\mu})^2,
\end{equation}
where $\hat{\mu}$ is the mean estimator over $n$ data points, $\hat{\mu}'$ is an extension of the mean adding $1/\tau^2$ virtual points, and $\hat{\sigma}^2$ is the estimator of the variance. Note that for the variance the biased version with $n$ was used instead of with $n-1$ as it allows to compute the Residual Sum of Squares (RSS) directly by $RSS = n \hat{\sigma}$.

Focusing now on the interior part of the integral of Eq.~\ref{eq:BayesStep3} and rewriting it in order to resemble the t-student distribution, we obtain: 
\begin{equation}
\begin{split}
&\left[ \sum_i^{n} (y_i-\mu)^2 +\frac{1}{\tau^2}(0-\mu^2) \right]^{-(n+1)/2} = \\
& \left[\sum_i^{n} y_i^2- (n+1)\hat{\mu}'^2 + (n+1)\hat{\mu}'^2 -2 (n+1/ \tau^2)\hat{\mu}'\mu^2 + (n+1/ \tau^2)\mu^2 \right]^{-(n+1)/2}=\\
& \left[\sum_i^{n} y_i^2- n\hat{\mu}^2 +(n+1/ \tau^2)(\hat{\mu}'-\mu)^2 \right]^{-(n+1)/2}=\\
&\left[ n \hat{\sigma}^2 +(n+1/\tau^2)(\hat{\mu}'-\mu)^2 \right]^{-(n+1)/2} =\\
&\left[n\hat{\sigma}^2\right]^{-(n+1)/2}  \left[1  + \frac{(n+1/\tau^2)(\hat{\mu}'-\mu)^2}{n\hat{\sigma}^2}  \right]^{-(n+1)/2} \\
&\left[n\hat{\sigma}^2\right]^{-(n+1)/2}  \left[1  + \frac{1}{n} \left( \frac{\hat{\mu}'-\mu}{s_s^2} \right)^2  \right]^{-(n+1)/2}, \\
\end{split}
\end{equation}
where $s_s^2 = \hat{\sigma}/(n+1/\tau^2)$ is the ``sampling" variance. Now, taking into account the fact that the integral of the t-student distribution over the whole space is equal to one, and reshuffling around its terms we get
\begin{equation}
\int_{-\infty}^{+\infty}  \left[1 + \frac{1}{n} \left( \frac{\hat{\mu}'-\mu}{s_s}  \right)^2   \right]^{-\frac{n+1}{2}} \dif \mu = \frac{\Gamma(\frac{n}{2})\sqrt{\pi n}s_s}{\Gamma(\frac{n+1}{2})}.
\end{equation}
Inserting this back in Eq.~\ref{eq:BayesStep2} we obtain:

\begin{equation}
\begin{split}
&P_{Bayes} (Y_n)= \\
&= \tau^{-1}\Gamma \left(\frac{n+1}{2} \right) 2^{\frac{n+1}{2}-1} (2 \pi)^{-\frac{n+1}{2}} \frac{\Gamma(\frac{n}{2})\sqrt{\pi n}s_s}{\Gamma(\frac{n+1}{2})} \left[n\hat{\sigma}^2\right]^{-(n+1)/2}  \\
&= \tau^{-1} 2^{-1}\pi^{-\frac{(n+1)}{2}} \Gamma(\frac{n}{2})\frac{1}{\sqrt{n+1/\tau^2}}   \left[n\hat{\sigma}^2\right]^{-\frac{n}{2}},  \\
\end{split}
\end{equation}

Returning to the the conditional probability of Equation~\eqref{eq:conditionalbayes}, we see that we still need to define $P(\Ytwon)$, the non-optimal probability of the first two-points. As in the case of our model class we assume that the dataset overall statistics are known, i.e., $\Theta = \{\hat{\mu}_d, \hat{\sigma}_d \}$, we will use this distribution to find the probability of the points $\Ytwon = \{y_1,y_2\}$ as :
\begin{equation}
P(\Ytwon) = \log 2 \pi + \log \hat{\sigma}_d +\left[ \frac{1}{2 \hat{\sigma}_d^2} \sum_i^2 (y_i-\hat{\mu}_d)^2   \right] \log e. \\
\end{equation}

Finally, applying the minus logarithm base $2$ to all the terms in Eq~\eqref{eq:conditionalbayes} to obtain the total code length in bits,
\begin{equation}\label{eq:BayesStepFinal}
\begin{split}
&L(Y_n) = -\log P_{Bayes} (Y_n) +\log P_{Bayes} (\Ytwon) - \log P(\Ytwon)  \\
&=1+ \frac{n}{2} \log \pi- \log \Gamma \left( \frac{n}{2} \right) + \frac{1}{2} \log ( n+1/\tau^2) +\frac{n}{2} \log \left (n\hat{\sigma}_n^2  \right)\\
&-1- \frac{2}{2}\log \pi+ 0 - \frac{1}{2} \log ( 2+1/\tau^2) -\log \left (\sum_i^2 (y_i-\hat{\mu}_2)^2   \right) \\
&+ \frac{2}{2}\log \pi+ \log \hat{\sigma}_d +\left[ \frac{1}{2 \hat{\sigma}_d^2} \sum_i^2 (y_i-\hat{\mu}_d)^2   \right] \log e  \\
& = \frac{n}{2} \log \pi- \log \Gamma \left( \frac{n}{2} \right) + \frac{1}{2} \log ( n+1/\tau^2) +\frac{n}{2} \log \left (n\hat{\sigma}_n^2  \right) + L_{cost}(\Ytwon),
\end{split}
\end{equation}

where $\hat{\mu}_2$ is the estimated mean of $y_1,y_2$ and $L_{cost}(\Ytwon)$ is the extra cost incurred of not being able to use a refined encoding for $\Ytwon$. Now that the length of the encoding is defined we just need to choose the two points. i.e., $y_1,y_2$. Because we want to minimize this length, we notice that there are only two terms that contribute to it in $L_{cost}(\Ytwon)$, and thus by choosing the two observations close to $\hat{\mu}_d$ we can both minimize the encoding of $P(\Ytwon)$ and maximize $P_{Bayes} (\Ytwon)$ for most cases. There are exceptions to this, depending on the respective values of $\mu_d$ and $y_1,y_2$ but this are not significant to change the values too much and also requires less computational search to find the points.
\section{Bayesian encoding convergence to BIC for large $n$}\label{appendix:BIC}
In this section it is shown that for large number of instances $n$ the Bayesian encoding of a normal distribution with unknown mean and standard deviation (Eq.~\eqref{eq:BayesStepFinal}) converges to the encoding of a normal distribution with mean and standard deviation known plus $\log n$, i.e., proportional to the definition of the Bayes Information Criterion (BIC).
First the encoding of a normal distribution with mean and standard deviation known over $n$ \emph{i.i.d.} points is equal to the sum of the individual encodings: 
\begin{equation}
L(Y \given \hat{\Theta}) = \frac{n}{2} \log 2\pi + \frac{n}{2} \log \hat{\sigma}^2 +  \left[ \frac{1}{2 \hat{\sigma}^2} \sum_i^n (y_i-\hat{\mu})^2   \right] \log e. \\
\end{equation}
Second, we need to use the Stirling's approximation of the Gamma function for large $n$:
\begin{equation}
\begin{split}
&- \log \Gamma \left( \frac{n}{2} \right)  \\
&\sim -\frac{1}{2}\log \pi  -\frac{1}{2}\log \left (n-2 \right) - \left (\frac{n}{2}-1 \right) \log \left (\frac{n}{2}-1 \right) + \left (\frac{n}{2}-1 \right) \log e, \\
\end{split}
\end{equation}
and finally we insert it into Eq.~\eqref{eq:BayesStepFinal} and assume $\tau = 1$ to obtain:
\begin{equation}
\begin{split}
&L(Y_n)  \sim \\
&\sim 1 +\frac{n-1}{2} \log \pi + \frac{1}{2} \log \left (\frac{n+1}{n-2} \right) +\frac{n}{2} \log \left ( \frac{n\hat{\sigma}^2}{n/2-1} \right)+ \left (\frac{n}{2}-1\right) \log e  \\ 
&+\log \left (\frac{n}{2}-1 \right) + L_{cost}(\Ytwon)  \\
& \sim \frac{n}{2} \log \pi + \frac{n}{2} \log 2\hat{\sigma}^2 + \left[ \frac{1}{2 \hat{\sigma}^2} \sum_i^n (y_i-\mu)^2   \right] \log e  + \log n -\log e + L_{cost}(\Ytwon)\\
&= L(Y \given \hat{\Theta}) + \log \frac{n}{e}  + L_{cost}(\Ytwon)\\
& \sim \frac{1}{2} \left( 2L(Y \given \hat{\Theta}) + 2\log n -2\log e \right) \\
&= \frac{1}{2}BIC,
\end{split}
\end{equation}
where from the second to the third line we assumed large $n$, making some of the terms disappear, while the definition $n \hat{\sigma}^2 = \sum_i^n (y_i-\mu)^2 $ is used for making the third term of the third expression appear. From the fourth to the fifth expressions it was assumed that $L_{cost}(\Ytwon)$ is negligible, as it is the cost of not being able to encode the first two points optimally. For the Bayes information criterion we used its standard definition,
\begin{equation}
BIC = 2 \ln \hat{\mathcal{L}} + k\ln n,
\end{equation}
where $\hat{\mathcal{L}}$ is the likelihood as estimated from the data, and $k$ is the number of parameters, which in our case is $2$.\\

\section{Kullback-Leibler divergence between two normal distributions}\label{appendix:KLD}

Let us assume two normal probability distributions, $p(x) \sim \mathcal{N}(\mu_p,\sigma_p)$ and  $q(x) \sim \mathcal{N}(\mu_q,\sigma_q)$. The Kullback-Leibler divergence of $q$ from $p$ is:
\begin{equation}
\begin{split}
D_{KL}(p; q) &= \int_{-\infty }^{+\infty } p(x)\log p(x) \dif x - \int_{-\infty }^{+\infty } p(x)\log q(x) \dif x\\
&= \mathbb{E}_p \left [   \log p(x) \right ] - \mathbb{E}_p\left [   \log q(x) \right ] \\
&= -\frac{1}{2} \left(\log e+ \log 2\pi\sigma_p^2 \right)  +\frac{1}{2}\log 2\pi \sigma_q^2 +\mathbb{E}_p\left [ \frac{(x-\mu_q)^2}{2\sigma_q^2} \log e  \right ] \\
&= -\frac{\log e}{2} + \log  \frac{\sigma_p}{\sigma_q} +\mathbb{E}_p\left [  \frac{x^2-2x\mu_q+ \mu_q^2}{2\sigma_q^2} \log e  \right ] \\
&= -\frac{\log e}{2} + \log  \frac{\sigma_q}{\sigma_p} +\frac{\sigma_p^2+\mu_p^2-2\mu_p\mu_q+ \mu_q^2}{2\sigma_q^2} \log e  \\
&= -\frac{\log e}{2} + \log  \frac{\sigma_q}{\sigma_p} +\frac{\sigma_p^2+(\mu_2- \mu_q)^2}{2\sigma_q^2} \log e.  \\
\end{split}
\end{equation}

Note that in the specific case where the Kullback-Leibler divergency only takes into account the means and assumes both standard deviations equal, i.e., $p(x) \sim \mathcal{N}(\mu_p,\sigma)$ and  $q(x) \sim \mathcal{N}(\mu_q,\sigma)$ one obtains: 
\begin{equation}
D_{KL}(p; q) = \frac{(\mu_2- \mu_q)^2}{2\sigma^2} \log e,
\end{equation}

and the weighted version of this $D_{KL}$, i.e., $WKL_{\mu} = n D_{KL}(p; q)$, is similar to the most common subgroup discovery quality functions used for numeric targets that do not take into account the dispersion of the subgroup, such as the weighted relative accuracy or the mean-test \cite{van2012diverse}, which is the square root of $WKL_{\mu}$. We will call this measure the Weighted Kullback-Leibler without dispersion.

\section{Equivalence between MDL-based subgroup lists model class and subgroup discovery quality measures}\label{appendix:WKL}

In this section we show that minimizing the MDL score in the case of a subgroup list of size $1$, i.e., it only contains one subgroup, is equivalent to maximizing the weighted Kullback-Leibler diverngence---a subgroup discovery quality measure \cite{van2010maximal}. 
First, we note that the selected model should be the one that minimizes the MDL score according to
\begin{equation}\label{eq:LengthTotal2}
M^* = \argmin_{M \in \M}  L(Y \given X,M)  + L(M),
\end{equation}
where all models $M$ in the model class $\mathcal{M}$ are composed of a subgroup $s$---with antecedent $a$ and its rule $r_a$---and of the default rule, $r_d$ with a distribution estimated over the whole dataset. The statistics associated with $r_a$ and $r_d $ are $\hat{\Theta}_a = \{ \hat{\mu}_a, \hat{\sigma}_a\}$ and $\hat{\Theta}_d = \{ \hat{\mu}_d, \hat{\sigma}_d\}$, respectively. $s$ is only activated in the subset $D_a = \{X_a,Y_a \}$ where is description is true, and $r_d$ is only active over the part of the dataset where $s$ is not present, i.e., $D_d = \{X_d,Y_d \} = \{X_{\neg a}, Y_{\neg a} \}$.

Focusing on $L(Y \given X,M)$ we see that its encoding is equal to:
\begin{equation}
\begin{split}
L(Y \given X,M) & = L(Y_a \given X_a) + L(Y_{\neg a} \given X_{\neg a}, \hat{\Theta}_d).
\end{split}
\end{equation}

Using the approximation derived in the previous sections for large $n$  (Appendix~\ref{appendix:BIC}) the encoding of the subgroup equals:
\begin{equation}
L(Y_a \given X_a) \sim \frac{n_a}{2} \log 2\pi + \frac{n_a}{2} \log \hat{\sigma}_a^2 +  \left[ \frac{1}{2 \hat{\sigma}_a^2} \sum_{y_i \in Y_a} (y_i-\hat{\mu}_a)^2   \right] \log e  + \log \frac{n_a}{e}
\end{equation}
and the encoding length of the default rule $r_d$ is equal to:
\begin{equation}
L(Y^{d} \given X^{d}, \hat{\Theta}_d) = \frac{n_d}{2} \log 2\pi + \frac{n_d}{2} \log \hat{\sigma}_d^2 +  \left[ \frac{1}{2 \hat{\sigma}_d^2} \sum_{y_i \in Y_d} (y_i-\hat{\mu}_d)^2   \right] \log e.
\end{equation}

Turning the problem into a maximization problem by multiplying by minus one and adding a constant $L(Y \given  \hat{\Theta}_d)$---the encoded size of the whole target $Y$ using the overall distribution statistics $\hat{\Theta}_d$--- we obtain:
\begin{equation}\label{eq:MDLsubgroup}
\begin{split}
&L(Y \given  \hat{\Theta}_d)-L(Y \given X,M)  \\
& =  L(Y_a \given  \hat{\Theta}_d)+\cancel{L(Y_{\neg a} \given  \hat{\Theta}_d)}-L(Y_a \given X_a) -\cancel{L(Y_{\neg a} \given X_{\neg a}, \hat{\Theta}_d)} \\
& = \frac{n_a}{2} \log \frac{\hat{\sigma}_d^2}{\hat{\sigma}_a^2} +\left[ \frac{1}{2 \hat{\sigma}_d^2} \sum_{y_i \in Y_a} (y_i-\hat{\mu}_d)^2   \right] \log e - \frac{n_a}{2} \log e - \log n\\
& = \frac{n_a}{2} \log \frac{\hat{\sigma}_d^2}{\hat{\sigma}_a^2} +\left[ \frac{ \sum_{y_i \in Y_a} y_i^2 - n \hat{\mu}_a^2 + n \hat{\mu}_a^2 -2n \hat{\mu}_a\hat{\mu}_d-\hat{\mu}_d)^2}{2 \hat{\sigma}_d^2}  \right] \log e \\
&\phantom{=,}- \frac{n_a}{2} \log e - \log n \\
& = \frac{n_a}{2} \log \frac{\hat{\sigma}_d^2}{\hat{\sigma}_a^2} +\frac{n_a}{2} \left[ \frac{\hat{\sigma}_a^2 +(\hat{\mu}_a -\hat{\mu}_d^2)^2}{\hat{\sigma}_d^2}  \right] \log e - \frac{n_a}{2} \log e - \log n \\
& = n_a D_{KL}(\hat{\Theta}_a; \hat{\Theta}_d) - \log n
\end{split}
\end{equation}
where $D_{KL}(\hat{\Theta}_a; \hat{\Theta}_d)$ represents the Kullback-Leibler divergence as defined in the previous section (Appendix~\ref{appendix:BIC}).
Finally, subtracting also the model encoding the following expression is obtained:
\begin{equation}\label{eq:MDLsubgroupmeasure}
L(Y \given \hat{\Theta}_d)-L(D,M) = n_a D_{KL}(\hat{\Theta}_a; \hat{\Theta}_d) - \log n - L(M)
\end{equation}
where the multiplication of $D_{KL}$ by the number of instances gives the weighted Kullback-Leibler Divergency, a subgroup discovery measure first introduced for nominal targets~\cite{van2010maximal}. The rest of the terms penalize multiple hypothesis testing, such as the variable term in $L(M)$, which penalizes the number of ways in which the dataset can be divided according to that variable. As an example. if we have a binary variable $x_1$, there are $2$ ways to divide the dataset, i.e., $x_1= True$ or $x_1= False$ and the term $L(v) = \log 2$ uniformly penalizes the fact that the variable has $2$ times (when compared with $x_1=True$ alone) of being correlated by chance. Similarly for categorical and numeric variables.

\section{Difference between rules for regression and subgroup discovery}\label{appendix:proof_sdvsregression}

In this section we show that minimizing the MDL score in the case of a regression problem for a rule list of size $1$ (without loss of generality for greater sizes), i.e., it only contains one rule, is different than maximizing a subgroup discovery quality measure such as the weighted Kullback-Leibler diverngence of Equation~\eqref{eq:MDLsubgroupmeasure}. First, we need to define a model class $\mathcal{M}$ for regression rule lists of size $1$, following the steps of rule lists for classification \cite{proencca2020interpretable}, of the form:

\begin{equation}\label{eq:modelclass2} \small
\begin{split}
\text{rule 1}: & \textsc{ if }   a \sqsubseteq  \x   \textsc{ then }  \hat{f}_{a,\hat{\mu},\hat{\sigma}}(y)\\ 
\text{default}: &\textsc{ else }      \hat{f}_{\neg a,\hat{\mu},\hat{\sigma}}(y)
\end{split}
\end{equation}

where the first rule is defined by the parameter set $\Theta_a = \{\hat{\mu}_a, \hat{\sigma}_a \}$ and the default rule by $\Theta_{\neg a} = \{\hat{\mu}_{\neg a}, \hat{\sigma}_{\neg a} \}$. Contrary to our definition of a subgroup list, the default rule is not fixed and varies depending on the first rule. There are many definitions of rule lists that use a fixed rule, however having a variable default rule that maximizes the prediction quality is the best representative of rule lists and of the objective of finding the best machine learning model, i.e., returning the best partition of the data with the smallest error possible. Note that a decision tree is also part of this family of models as any path starting at the root of the tree to one of its leaves also forms a rule, and thus, a decision tree is equivalent to a set of disjoint rules, i.e., none of the rules described in this way overlap on a dataset. For this type of regression rule lists, the encoding of the first rule and default rule is given by Equation~\ref{eq:BayesStepFinal} as in both cases the parameters are unknown.

Thus the data encoding of the regression rule list is given by:
\begin{equation}
\begin{split}
L(Y \given X,M) & = L(Y_a \given X_a) + L(Y_{\neg a} \given X_{\neg a}),
\end{split}
\end{equation}
and the model encoding $L(M)$ has the same form as for a subgroup list.

Following the same steps as in the Appendix~\ref{appendix:WKL} as multiplying by minus one to make it a maximization problem and adding $L(Y \given  \hat{\Theta}_d)$:

\begin{equation}\label{eq:MDLregression}
\begin{split}
&L(Y \given  \hat{\Theta}_d)-L(Y \given X,M)  \\
& =  L(Y_a \given  \hat{\Theta}_d)+L(Y_{\neg a} \given  \hat{\Theta}_d)-L(Y_a \given X_a) -L(Y_{\neg a} \given X_{\neg a} ) \\
& = n_a D_{KL}(\hat{\Theta}_a; \hat{\Theta}_d) - \log n_a +  n_{\neg a} D_{KL}(\hat{\Theta}_{\neg a}; \hat{\Theta}_d) - \log n_{\neg a},\\
\end{split}
\end{equation}

and if one strips the model complexity part, one arrives at a ``quality'' measure of $ n_a D_{KL}(\hat{\Theta}_a; \hat{\Theta}_d)+ n_{\neg a} D_{KL}(\hat{\Theta}_{\neg a}; \hat{\Theta}_d) $.
Comparing both Eq.~\ref{eq:MDLsubgroup} and Eq.~\ref{eq:MDLregression} we can notice the most important distinction between subgroup discovery and regression: the \emph{local} nature of subgroup discovery and the \emph{global} nature of the regression task. In other words, subgroup discovery aims at finding rules that locally maximize their quality, independently of the rest of the dataset, and even though rules for regression try to maximize their \emph{local} quality also they have to take into account the quality of their negative set, i.e., a rule for regression cannot be considered by its quality alone, it has to be considered in terms of its \emph{global} impact in the dataset. 
On the other hand, this result also shows the similarity between both tasks and where the confusion sometimes arises, i.e., in some particular cases the best subgroup can be also the best regression rule. An example of this would be a dataset that is very large (relatively to the number of observations covered by the rule), and the best rule/subgroup would cover a small number of observations compared to the rule formed by the negative set of that rule, i.e., $D_{\neg a}$, as a similar distribution to $\Theta_d = \{\hat{\mu}_d, \hat{\sigma}_d \}$, making $\Theta_{\neg a} \sim \Theta_d$. Nonetheless, this similarity decreases in the case of larger lists, as the default rule for regression will always represent what is left and in a subgroup list it remains constant and representing what we consider uninteresting.
\pagebreak

\section{Empirical analysis of absolute versus normalized gain}\label{appendix:empiricalabsvsnorm}
\begin{figure}[!h]
	\centering
		\includegraphics[width=1.0\textwidth]{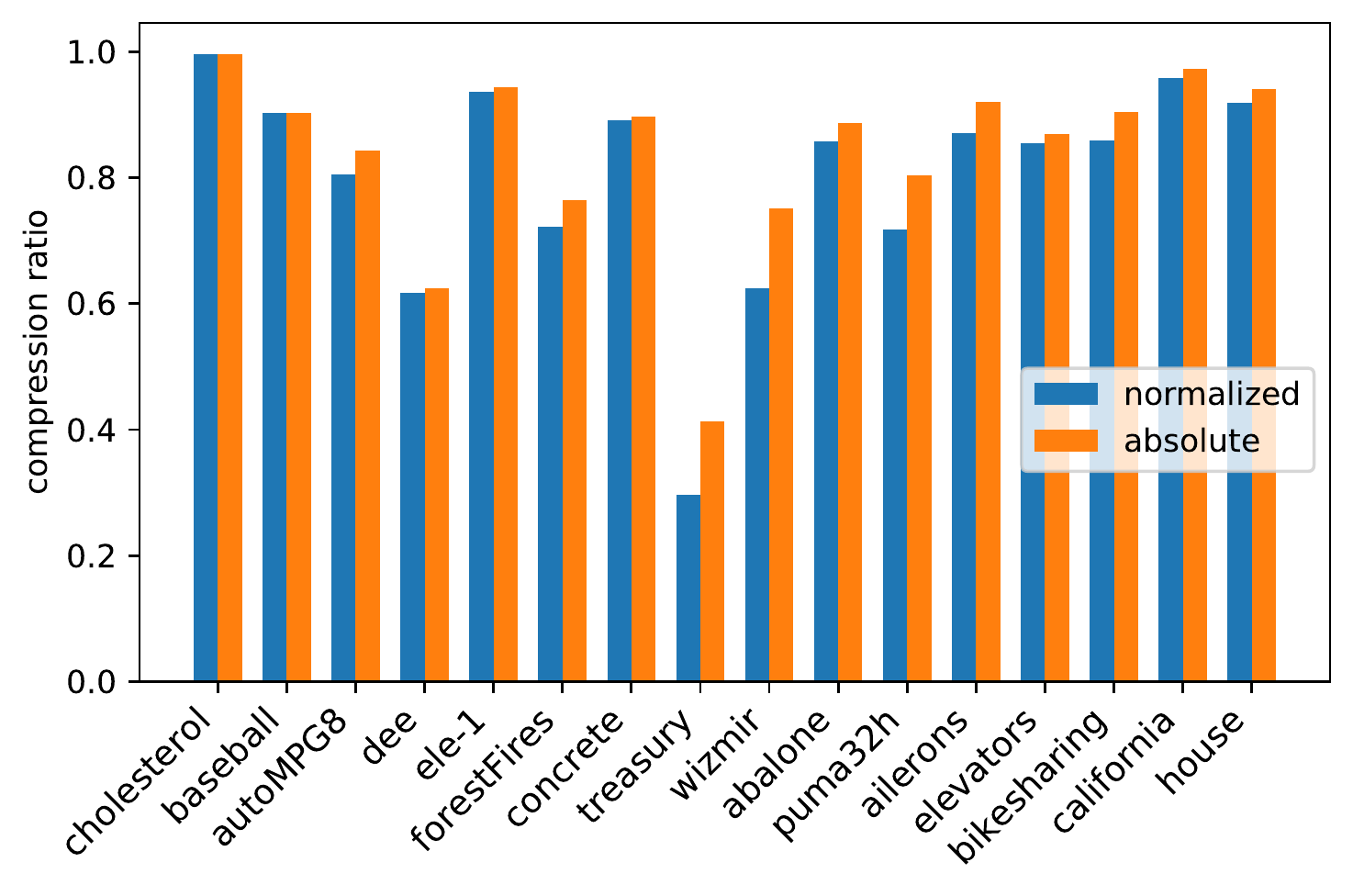} 
		\caption{Compression ratio obtained with normalized and absolute gain.}\label{fig:gain_compression}
\end{figure}

\begin{figure}[!h]
	\centering
		\includegraphics[width=1.0\textwidth]{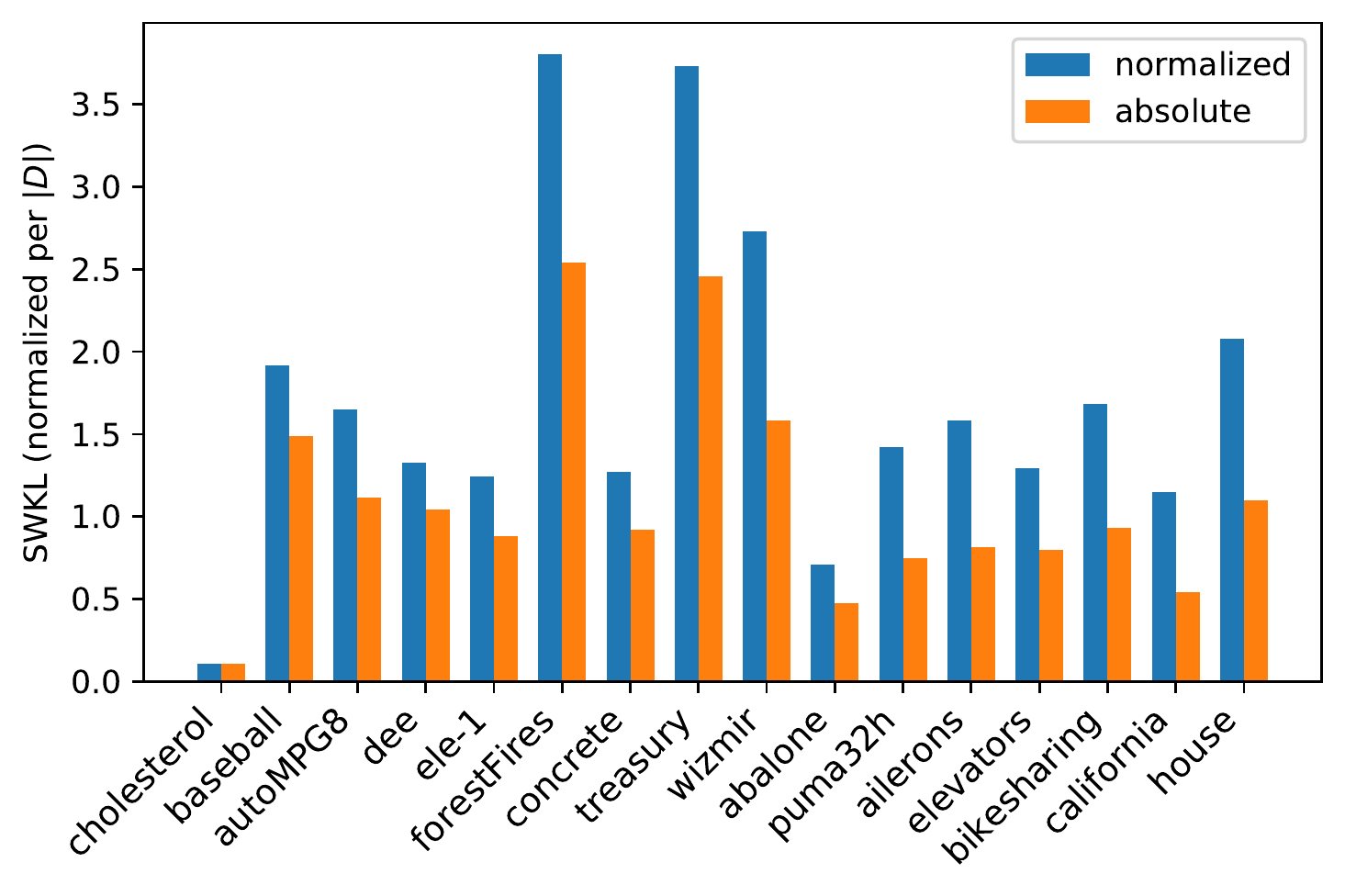} 
		\caption{Sum of Weighted Kullback-Leibler (SWKL) normalized by the number of instances per dataset obtained with normalized and absolute gain.}\label{fig:gain_swkl}
\end{figure}

\begin{figure}[!h]
	\centering
		\includegraphics[width=1.0\textwidth]{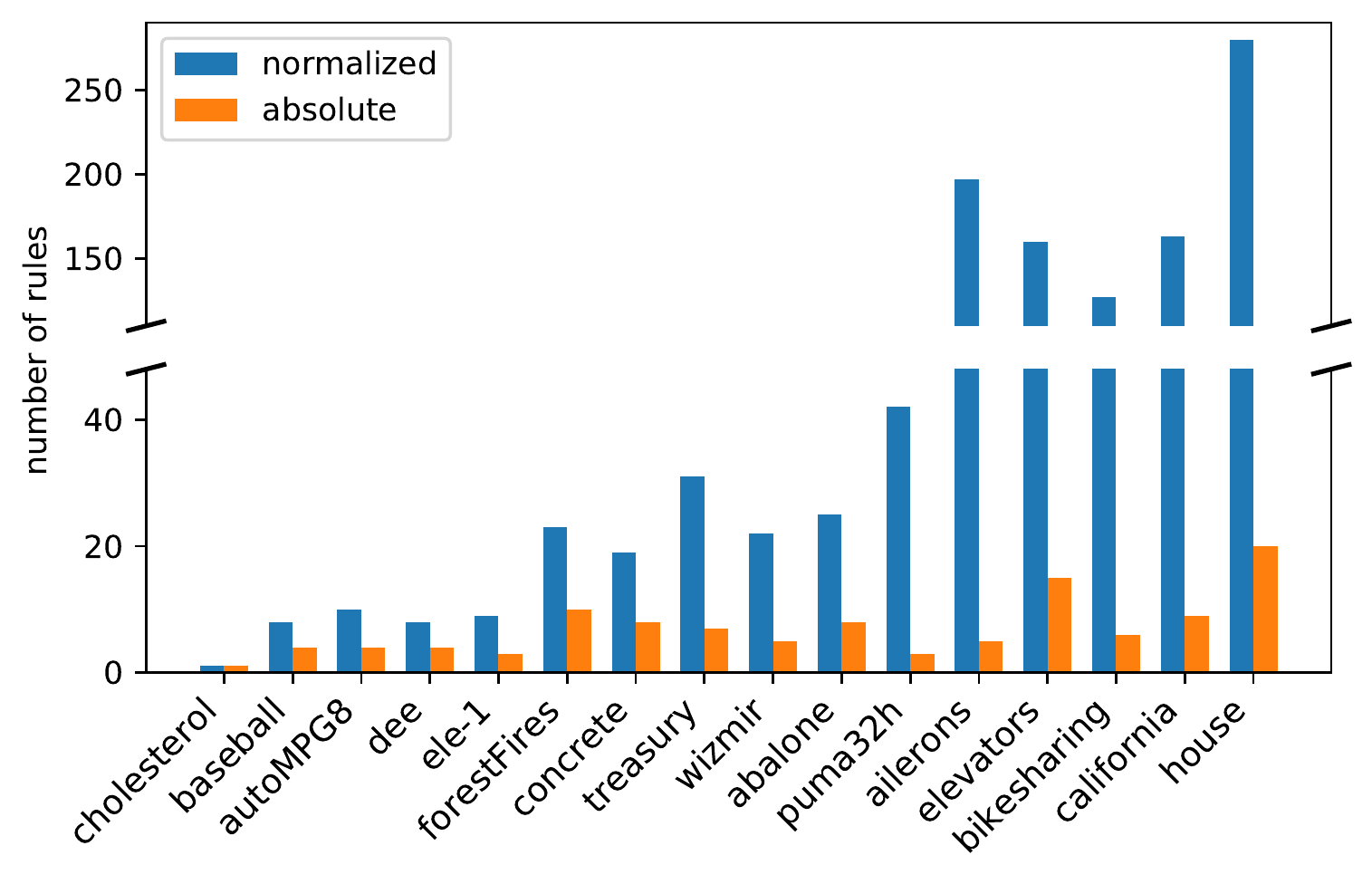} 
		\caption{Number of rules obtained with normalized and absolute gain.}\label{fig:gain_nrules}
\end{figure}

\begin{figure}[!h]
	\centering
		\includegraphics[width=1.0\textwidth]{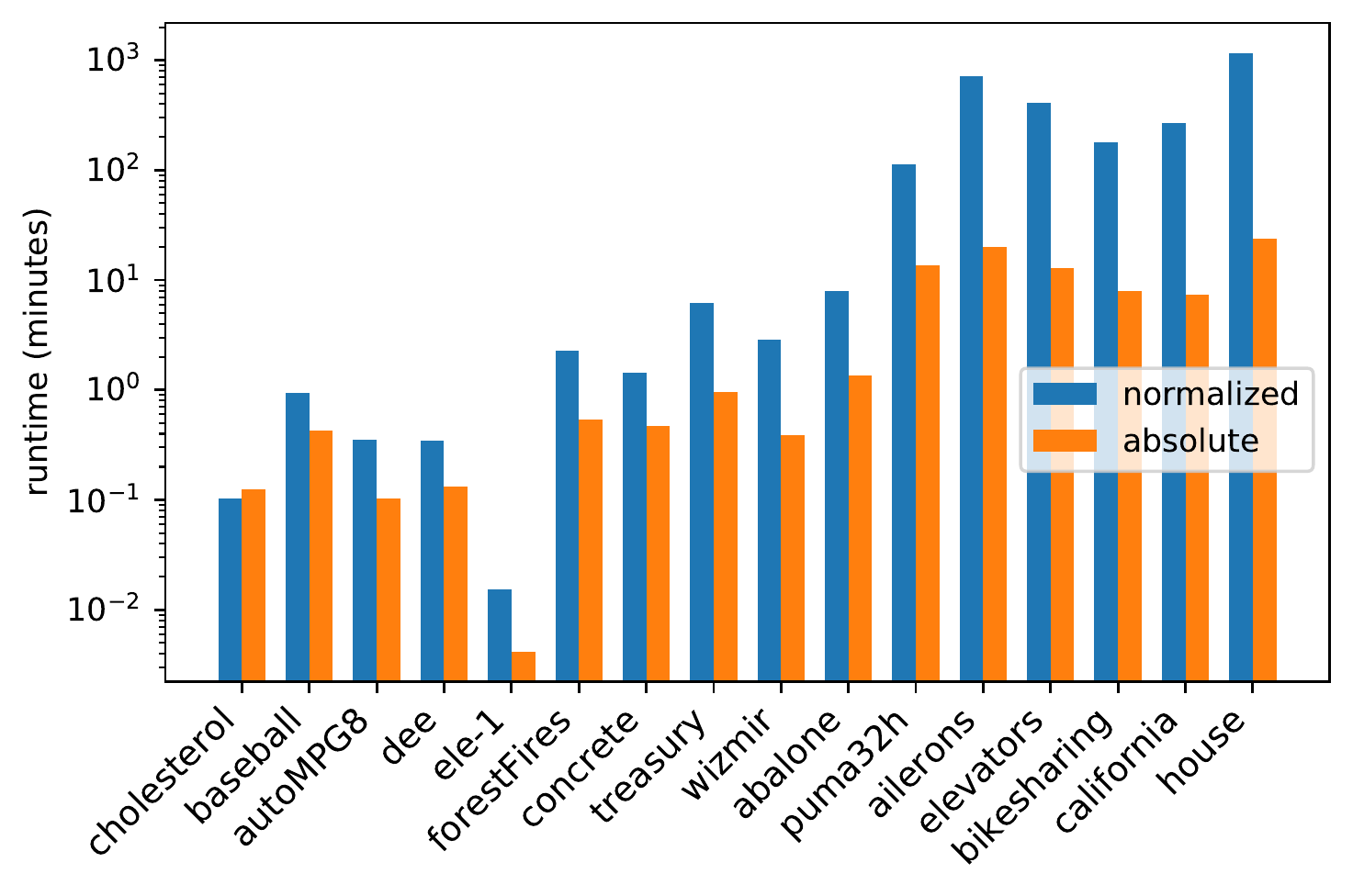} 
		\caption{runtime obtained with normalized and absolute gain.}\label{fig:gain_runtime}
\end{figure}

In this section we present a thorough comparisong of the SSD++ with \emph{absolute} gain and \emph{normalized} gain. SSD++ is executed with the same parameters (beam width, number of cut points for numerical variables, and maximum depth of search) as in the experiments section of the paper, i.e., $w_b =100$, $n_{cut} =5$, $d_{max} = 5$. Both types of gain are compared for all the benchmark datasets described in the paper in terms of their compression ratio (defined later) in Figure~\ref{fig:gain_compression}, Sum of Weighted Kullback-Leibler divergency (SWKL) in Figure~\ref{fig:gain_swkl}, number of rules in Figure~\ref{fig:gain_nrules}, and runtime in minutes in Figure~\ref{fig:gain_runtime}. The compression ratio is the length of the found model $L(D, M)$ divided by the length of just encoding the data with the dataset distribution (a model without subroups) $L(D \given \Theta_d)$, and formally as the following form:
\begin{equation}
L \% = \frac{L(D, M)}{L(D \given \Theta_d)}
\end{equation}


\pagebreak

\section{Empirical analysis of the influence of the beam search hyperparameters}\label{appendix:empiricalbeamsearch}
In this section we present a thorough comparison of the influence the change in the hyperparameters of the beam search of SSD++ on its results. As a complete search over the whole combination of parameters is unfeasible we present here a exploration over the parameters used for the experimental comparison in the paper ($w_b =100$, $n_{cut} =5$, $d_{max} = 5$), i.e., we fix two of the parameters on the aforementioned values and then proceed to change the selected parameter of interest, and then we do this for all the $3$ parameters. The line between the dots of the same color does not represent an interpolation and is merely used to aid visualization and suggest trends. 

\begin{figure}[!h]
	\centering
		\includegraphics[width=1.0\textwidth]{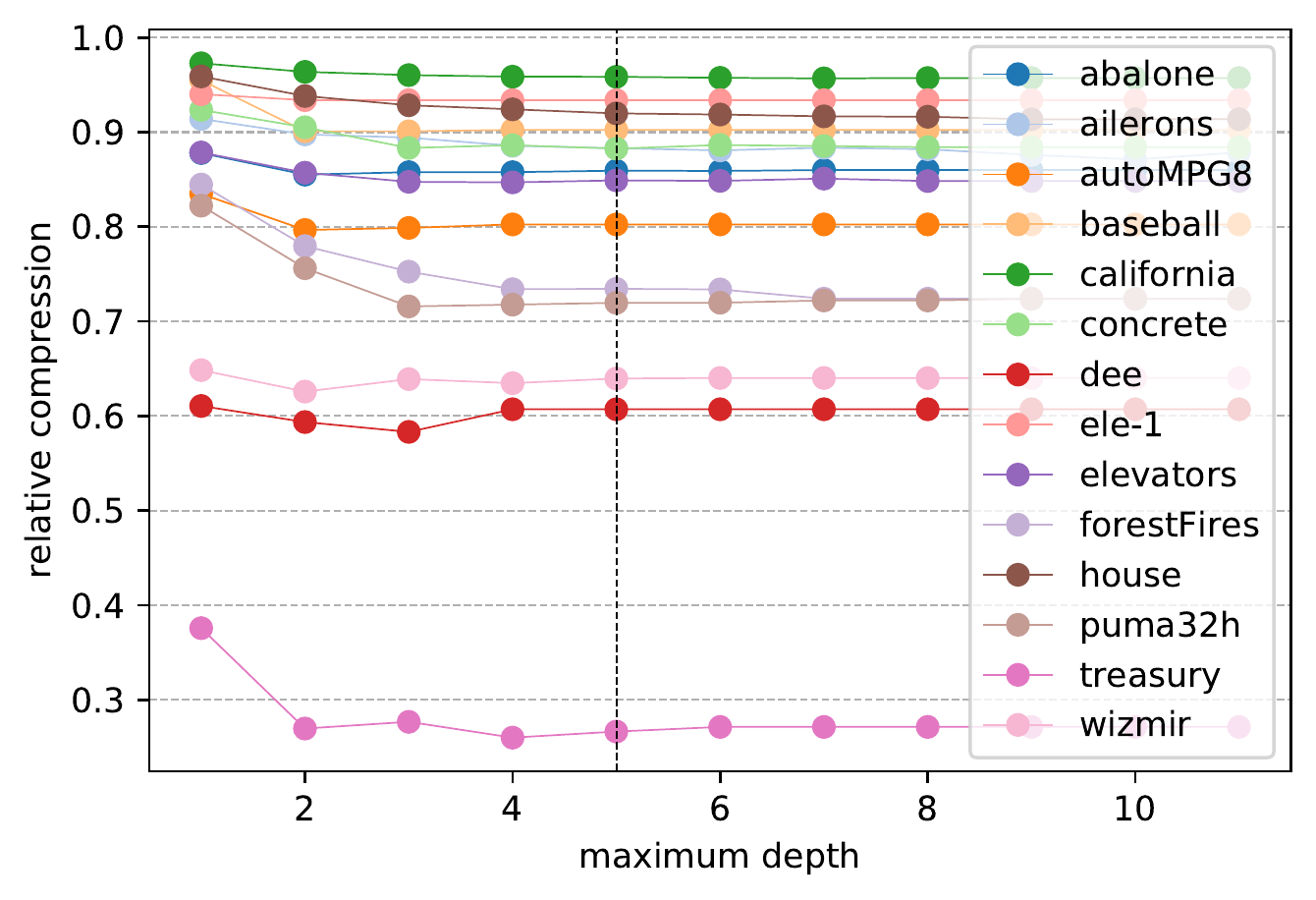} 
		\caption{Compression ratio obtained by varying the maximum search depth and fixing $w_b =100$, $n_{cut} =5$. The black vertical line represents the value used in Experiments section of the paper.}
\end{figure}
\begin{figure}[!h]
	\centering
		\includegraphics[width=1.0\textwidth]{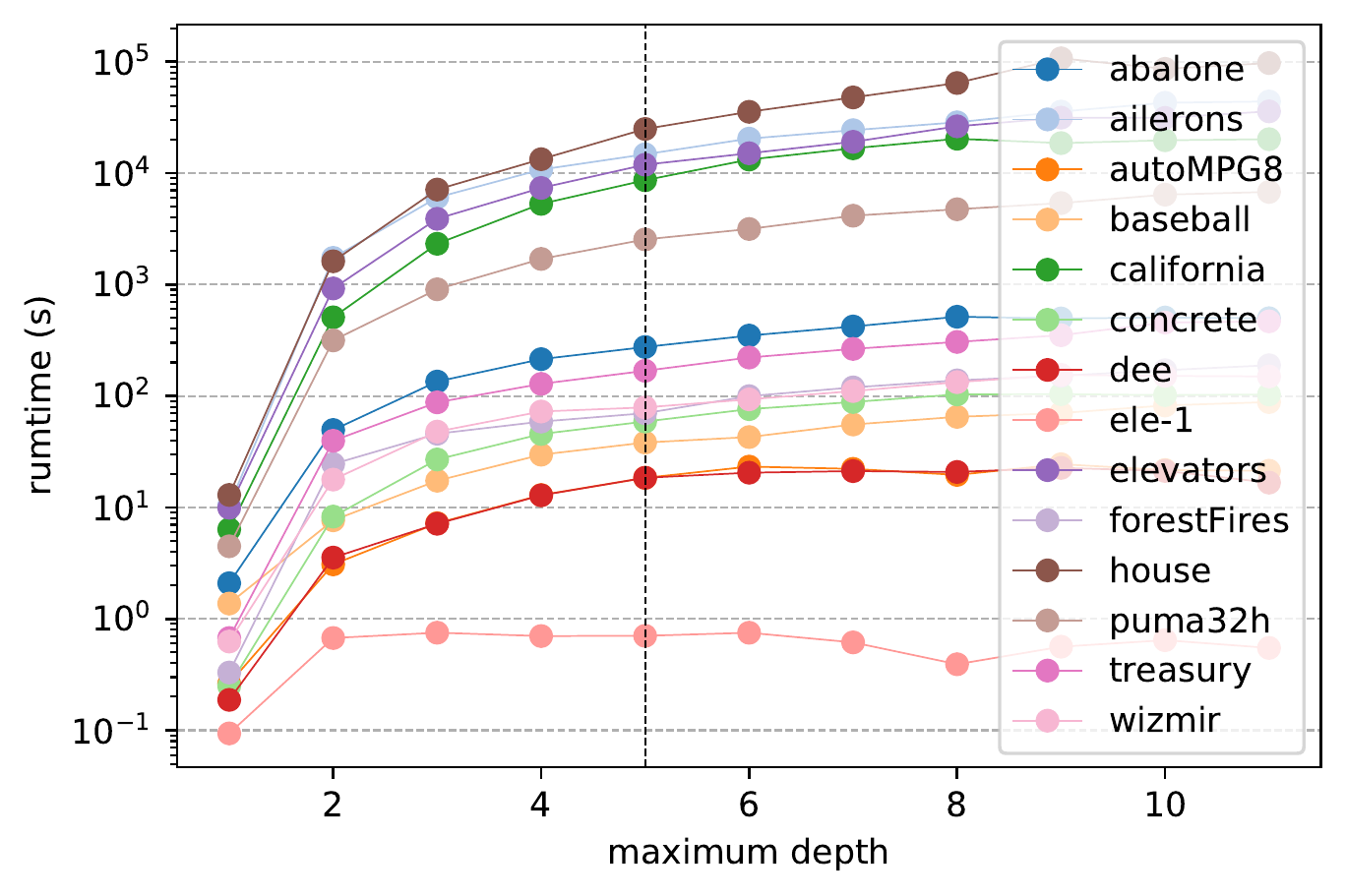} 
		\caption{Runtime in seconds obtained by varying the maximum search depth and fixing $w_b =100$, $n_{cut} =5$. The black vertical line represents the value used in Experiments section of the paper.}
\end{figure}
\begin{figure}[!h]
	\centering
		\includegraphics[width=1.0\textwidth]{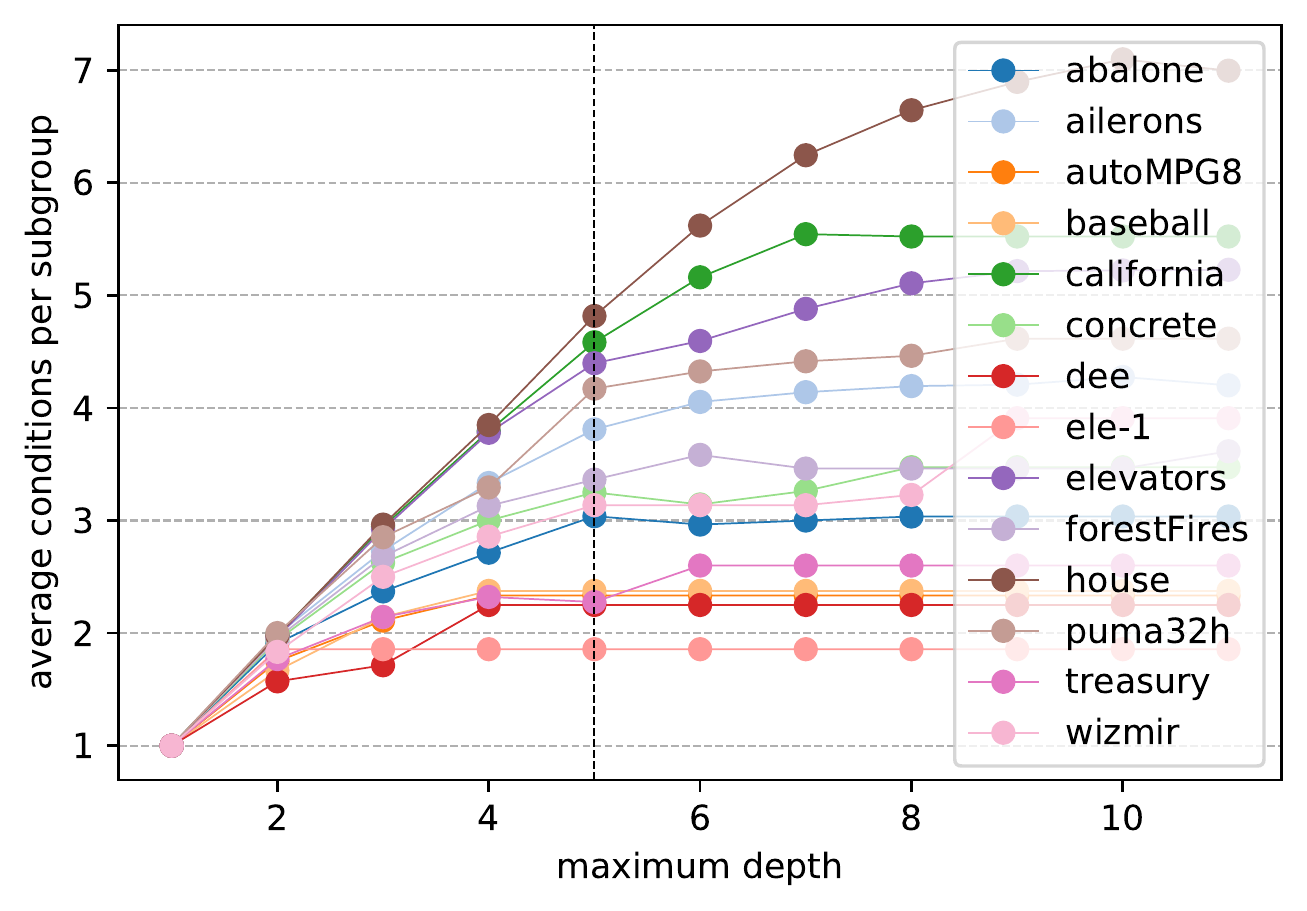} 
		\caption{Average number of conditions per subgroup obtained by varying the maximum search depth and fixing $w_b =100$, $n_{cut} =5$. The black vertical line represents the value used in Experiments section of the paper.}
\end{figure}
\begin{figure}[!h]
	\centering
		\includegraphics[width=1.0\textwidth]{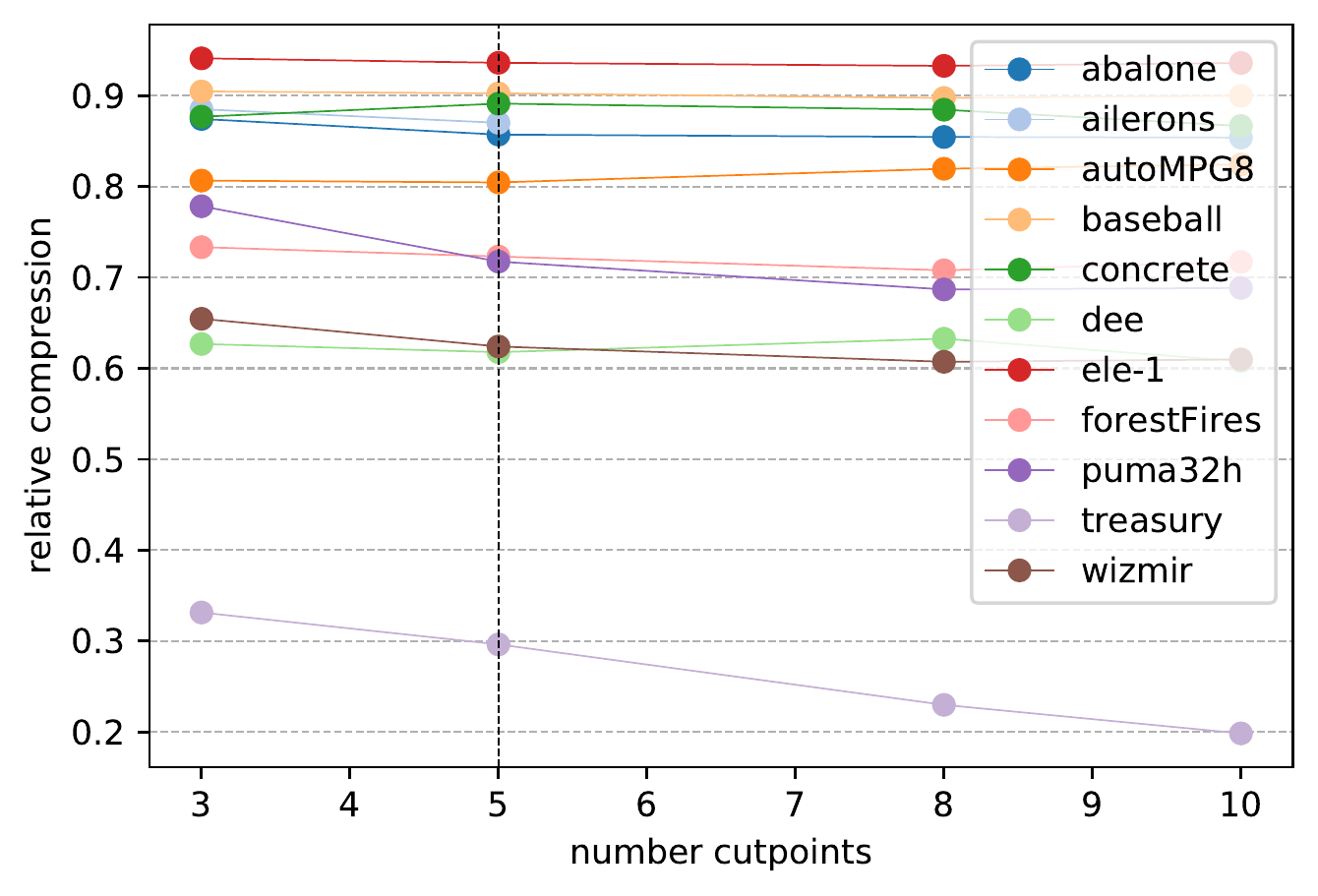} 
		\caption{Compression ratio obtained by varying the number of cut points and fixing $w_b =100$, $d_{max} =5$. The black vertical line represents the value used in Experiments section of the paper.}
\end{figure}												
\begin{figure}[!h]
	\centering
		\includegraphics[width=1.0\textwidth]{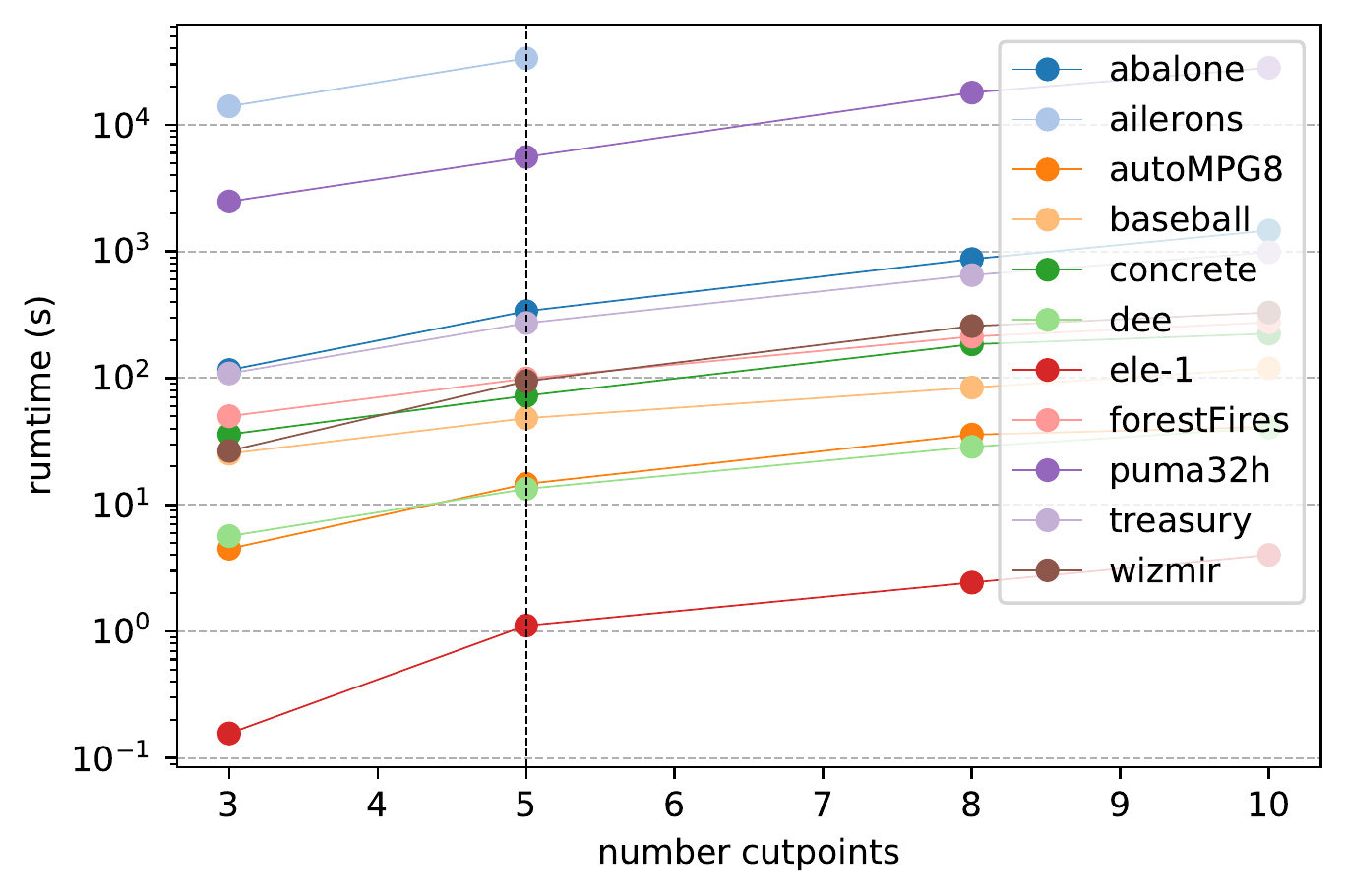} 
		\caption{Runtime in seconds obtained by varying the number of cut points and fixing $w_b =100$, $d_{max} =5$. The black vertical line represents the value used in Experiments section of the paper.}
\end{figure}	
\begin{figure}[!h]
	\centering
		\includegraphics[width=1.0\textwidth]{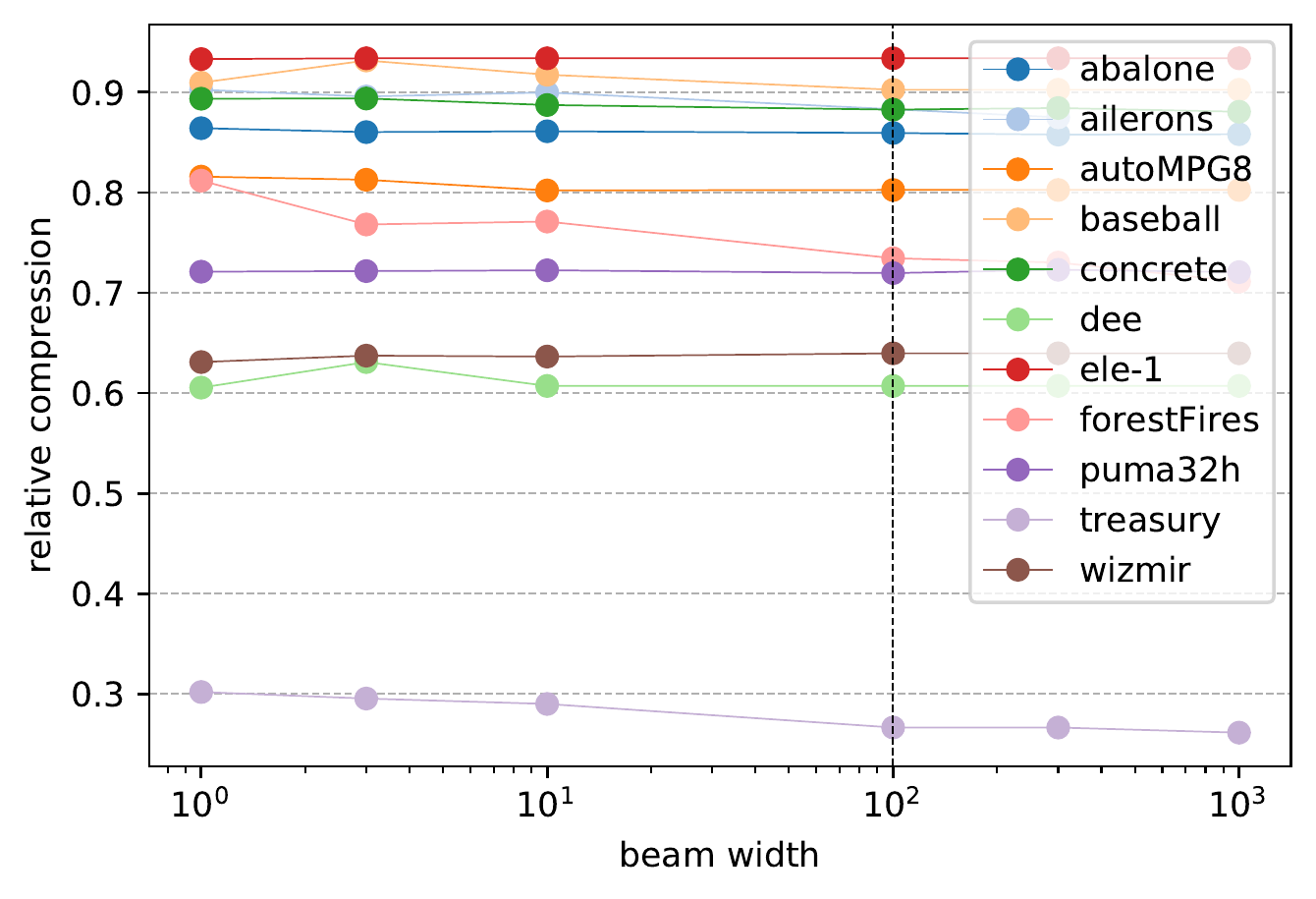} 
		\caption{Compression ratio obtained by varying the beam width and fixing $n_{cut} =5$, $d_{max} =5$. The black vertical line represents the value used in Experiments section of the paper.}
\end{figure}
\begin{figure}[!h]
	\centering
		\includegraphics[width=1.0\textwidth]{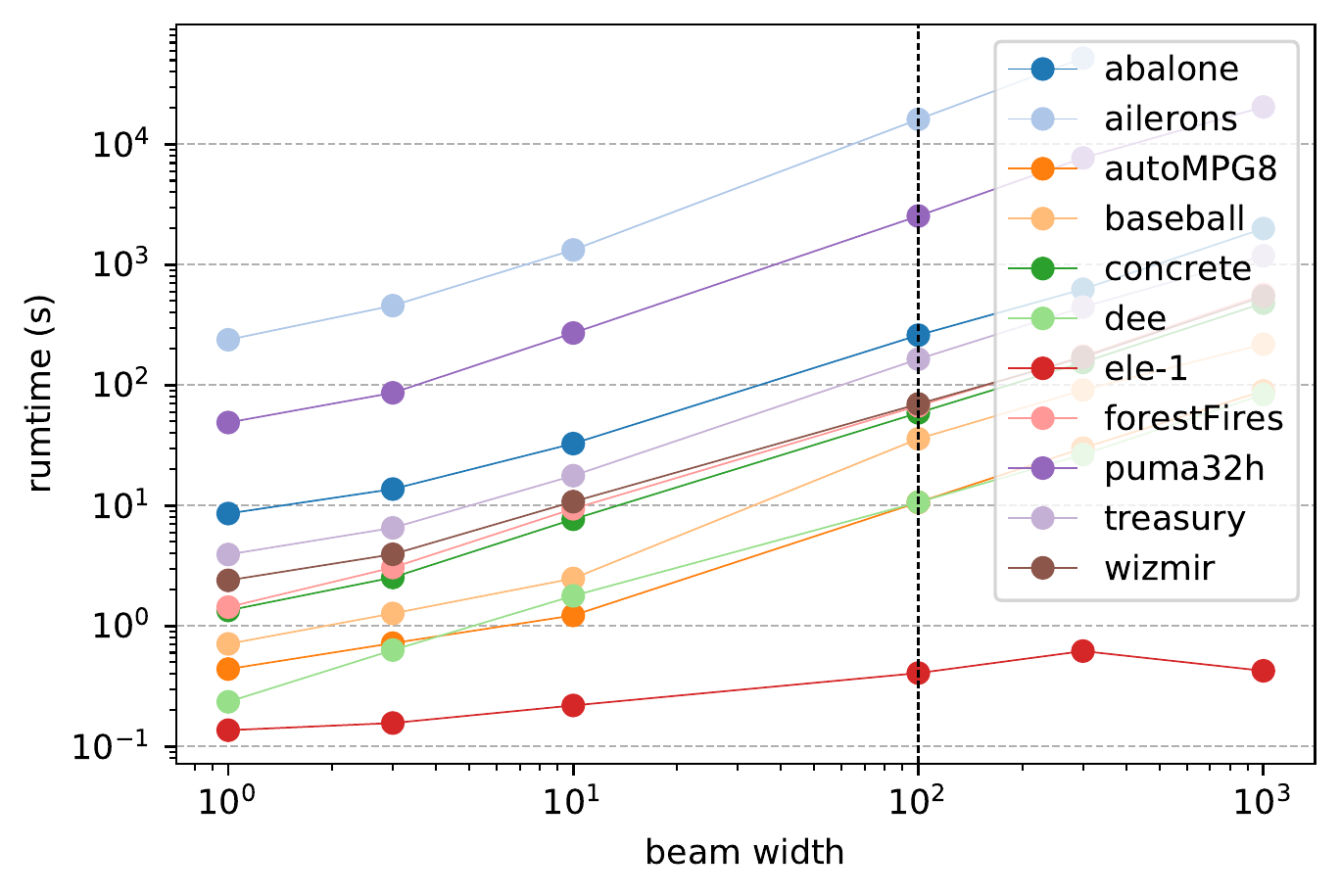} 
		\caption{Runtime in seconds obtained by varying the beam width and fixing $n_{cut} =5$, $d_{max} =5$. The black vertical line represents the value used in Experiments section of the paper.}
\end{figure}

\end{appendix}
\end{document}